\documentclass{article}

\usepackage[sort,authoryear,round]{natbib}
\usepackage{fullpage}

\usepackage[utf8]{inputenc} 
\usepackage[T1]{fontenc}    
\usepackage{xcolor}
\usepackage{hyperref}       
\definecolor{myblue}{rgb}{0,0.2,0.8}
\hypersetup{ %
pdftitle={},
pdfauthor={},
pdfsubject={},
pdfkeywords={},
pdfborder=0 0 0,
pdfpagemode=UseNone,
colorlinks=true,
linkcolor=myblue,
citecolor=myblue,
filecolor=myblue,
urlcolor=myblue,
pdfview=FitH}
\usepackage{authblk}
\usepackage{url}            
\usepackage{booktabs}       
\usepackage{amsfonts}       
\usepackage{amsthm}
\usepackage{nicefrac}       
\usepackage{microtype}      
\usepackage{amsmath} 

\usepackage{graphicx}
\usepackage{xspace}
\usepackage[varqu,varl,var0,scaled=0.97]{inconsolata}
\usepackage{caption}
\usepackage{overpic}
\usepackage{wrapfig}
\usepackage{tcolorbox}
\usepackage{enumitem}
\usepackage{threeparttable}
\usepackage{subfig}

\usepackage{listings}
\usepackage{color}
\usepackage{lipsum}

\definecolor{dkgreen}{rgb}{0,0.6,0}
\definecolor{gray}{rgb}{0.5,0.5,0.5}
\definecolor{mauve}{rgb}{0.58,0,0.82}

\usepackage[capitalize,noabbrev]{cleveref}

\newtheorem{theorem}{Theorem}

\newcommand\fakeparagraph[1]{\par\noindent\textbf{{#1}}.\xspace}

\newcommand{\figref}[1]{Fig.~\ref{#1}}

\newcommand{\tabref}[1]{Table~\ref{#1}}

\newcommand{\secref}[1]{Sec.~\ref{#1}}
\newcommand{\equref}[1]{Eq.~(\ref{#1})}

\newcommand{\eg}{\emph{e.g.},\xspace}
\newcommand{\ie}{\emph{i.e.},\xspace}

\newcommand{\cf}{\emph{cf.}\xspace}
\newcommand{\impscore}{$I$\xspace}
\usepackage{xcolor,soul}

\newcommand\reds{\textsc{REDS}\xspace}

\newcommand\vit{\textsc{ViT}\xspace}
\newcommand\msa{\textsc{MSA}\xspace}
\newcommand\mlp{\textsc{MLP}\xspace}
\newcommand\llms{\textsc{LLMs}\xspace}
\newcommand\llm{\textsc{LLM}\xspace}

\newcommand\dnn{\textsc{DNN}\xspace}
\newcommand\cnn{\textsc{CNN}\xspace}
\newcommand\dscnn{\textsc{DS-CNN}\xspace}

\newcommand\bottomup{\textsc{BU}\xspace}
\newcommand\topdown{\textsc{TD}\xspace}

\newcommand\cifar{\textsc{CIFAR10}\xspace}
\newcommand\imagenet{\textsc{ImageNet-1K}\xspace}
\newcommand\fmnist{\textsc{Fashion-MNIST}\xspace}
\newcommand\mobilenet{\textsc{MobileNetV1}\xspace}
\newcommand\visualwake{\textsc{VWW}\xspace}
\newcommand\googlespeech{\textsc{GSC}\xspace}

\title{\bf{\reds: Resource-Efficient Deep Subnetworks\\for Dynamic Resource Constraints}}
\author[1]{Francesco Corti}
\author[2]{Balz Maag}
\author[3]{Joachim Schauer}
\author[4]{Ulrich Pferschy}
\author[1,5]{Olga Saukh}
\affil[1]{Graz University of Technology, Austria}
\affil[2]{ABB Research, Switzerland}
\affil[3]{University of Applied Sciences, FH Joanneum, Austria}
\affil[4]{University of Graz, Austria}
\affil[5]{CSH Vienna, Austria}
\affil[ ]{}
\affil[ ]{\texttt{\{francesco.corti,saukh\}@tugraz.at, balz.maag@ch.abb.com, joachim.schauer@fh-joanneum.at, ulrich.pferschy@uni-graz.at}}

\begin{document}
\date{}
\maketitle

\begin{abstract}
Deep learning models deployed on edge devices frequently encounter resource variability, which arises from fluctuating energy levels, timing constraints, or prioritization of other critical tasks within the system. State-of-the-art machine learning pipelines generate resource-agnostic models that are not capable to adapt at runtime. In this work, we introduce Resource-Efficient Deep Subnetworks (\reds) to tackle model adaptation to variable resources. In contrast to the state-of-the-art, \reds leverages structured sparsity constructively by exploiting permutation invariance of neurons, which allows for hardware-specific optimizations. Specifically, \reds achieves computational efficiency by (1) skipping sequential computational blocks identified by a novel iterative knapsack optimizer, and (2) taking advantage of data cache by re-arranging the order of operations in \reds computational graph. \reds supports conventional deep networks frequently deployed on the edge and provides computational benefits even for small and simple networks. We evaluate \reds on eight benchmark architectures trained on the Visual Wake Words, Google Speech Commands, Fashion-MNIST, CIFAR-10 and ImageNet-1K datasets, and test on four off-the-shelf mobile and embedded hardware platforms. We provide a theoretical result and empirical evidence demonstrating \reds' outstanding performance in terms of submodels' test set accuracy, and demonstrate an adaptation time in response to dynamic resource constraints of under 40$\mu$s, utilizing a  fully-connected network on Arduino Nano 33 BLE.
\end{abstract}

\section{Introduction}
  \label{sec:intro}

Data processing pipelines in edge devices increasingly rely on deep learning models to identify patterns and extract insights from multimodal IoT data. Examples include predictive maintenance in industrial automation, object identification and tracking in smart camera systems~\citep{qu2022dress}, activity and healthcare trackers in mobile~\citep{ravi2016deep}, wearable and hearable applications~\citep{MaagZST17Barton,sabry2022machine}. In all these systems, deep models are deployed and run along with other tasks, under constraints and priorities dictated by the current context and available resources, including storage, CPU time, energy and bandwidth. A large body of work explores different techniques to optimize and compress deep models without hurting accuracy and generalization abilities, while accelerating their execution in software~\citep{boehm2018optimizing} and in hardware~\citep{jouppi2017datacenter}. Model pruning and quantization~\citep{han2016deep} have become part of standard deep learning deployment pipelines, \eg TFLMicro~\citep{TFLite} and TensorRT~\citep{vanholder2016efficient}, to enable deep learning on severely constrained embedded hardware operated by low-power microcontrollers with only a few kB of RAM. Pruning covers a set of methods that reduce the model size by eliminating unimportant operations (weights, neurons, kernels) in the model~\citep{dai2018compressing,li2017pruning}. These methods date back to the optimal brain damage~\citep{lecun1989optimal} and the optimal brain surgeon~\citep{hassibi1993second}, which suggest to prune the weights based on the Hessians of the loss function. Recent pruning methods~\citep{han2015learning,entezari2019class,timpl2022understanding, corti2022studying} propose to prune network weights with a low magnitude or a low magnitude increase. \citep{li2017pruning} propose to prune channels in CNNs based on a filter weight norm, while \cite{hu2016network} use the average percentage of zeros in the output to prune unimportant channels.

One drawback of compile-time optimizations, \eg pruning, is that the resulting models are \emph{resource-agnostic}. They thus yield suboptimal performance in many interesting applications, where resource availability depends on different dynamically changing factors such as available energy, task priority and timing constraints. Another drawback of one-shot model compression techniques, is that these are applied to the \emph{whole model}, making exploration of different options for a resource-aware on-device model reconfiguration challenging. Both problems are described in detail below.

\fakeparagraph{Dynamic resource constraints}
Many interesting applications can make use of resource-aware deep models, \ie models that can adapt their execution to available computational resources and time constraints. For example, camera image processing by a drone or a car may depend on the respective speed~\citep{qu2022dress}. Processing high-value data can justify using more energy and computational time than when running regular environment scans~\citep{gherman2021poster}. A naive solution to address dynamic resource constraints is to store several independent deep models and switch between them as resource availability and task priorities change. The drawback of this approach is the increased memory consumption to store these independent models, which does not scale, and the overhead of switching between models at runtime, \eg by loading these from flash to RAM and reallocating the necessary tensors.

Several approaches have been proposed to adapt a model to dynamic resource constraints. These methods can be classified into two groups: those that implement early exit predictions at a cost of a reduced accuracy, \eg BudgetRNN~\citep{kannan2021budget} and ASVN~\citep{bambusi21}, and those that build a subnetwork structure using weight sharing, with each subnetwork being more efficient yet possibly less accurate, \eg Slimmable Subnets~\citep{yu2018slimmable}, DRESS~\citep{qu2022dress} and NestDNN~\citep{fang2018nestdnn}. This work falls into the latter category but leverages structured sparsity constructively to enable additional flexibility and hardware support on IoT devices.

\begin{figure*}[t]
    \centering
    \includegraphics[width=\textwidth]{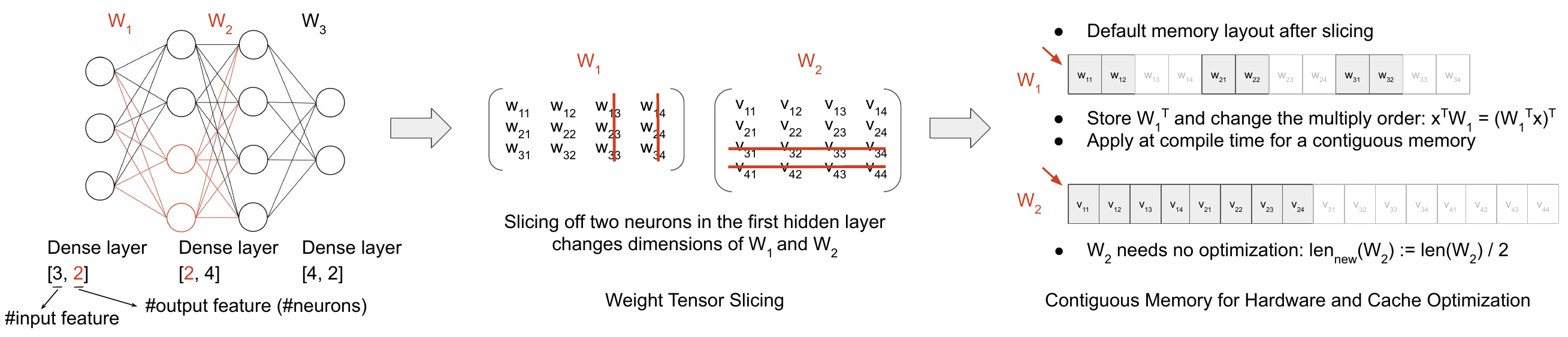}
    \caption{Example of layer slicing in a fully-connected model.  
    We slice off two neurons, \ie computational units, in the first hidden layer of the network with 4 neurons, 3 inputs and 2 outputs. The matrices $W_1$ and $W_2$ store the weights along all connections, and the respective columns and rows in $W_1$ and $W_2$ get eliminated by layer slicing. This breaks the contiguous memory layout and the memory arrangement of $W_1$, yet not $W_2$. A transpose of $W_1$ and a change of the multiply order preserve contiguous memory of $W_1$ after slicing.}
    \label{fig:REDS_Dense_splits}
\end{figure*}

\fakeparagraph{Deep learning support on IoT devices} 
Modern IoT hardware often provides support for running deep learning models. This support includes floating point instructions, dedicated instructions for frequent operations, such as MLA, FMA, and their vector versions. Deep neural network frameworks use hardware-specific features on supported platform to provide maximum speedup. However, the methods that introduce subnetwork structures to a model can not rely on the toolchain support to optimize the execution of each subnetwork. For example, fine-grained weight pruning may lead to accelerated execution due to an abundance of zeros in the weight matrices and sparse filters, yet unconstrained locations of these zeros present a difficulty in using unstructured sparsity. Additionally, specific algorithms are designed to effectively manage various sparsity levels, which may or may not be supported by the available hardware~\citep{Hoefler2021sparsity}. To overcome the issues, NestDNN~\citep{fang2018nestdnn} prunes convolutional filters; these have to be paged in or out when switching from one multi-capacity model to another. DRESS~\citep{qu2022dress} relies on specialized hardware to ensure the applied sparsity pattern translates into computation efficiency. Hardware accelerators may appear too power-hungry and expensive for battery-powered IoT devices. Specialized hardware is also inflexible as the deep learning technology advances fast and hardware modernizations are costly.  

\fakeparagraph{Contributions}
We present the design of Resource-Efficient Deep Subnetworks (\reds), featuring a nested submodel structure to address dynamic resource constraints on mobile and IoT devices. \reds introduces a novel hardware-aware method for designing nested subnetwork structures, \ie structured sparsity patterns, by formulating and solving an iterative knapsack problem.

This method provides both theoretical guarantees and empirical evidence of outstanding solution performance. Furthermore, we leverage permutation invariance of neurons~\citep{entezari2021role}  to keep the subnetwork weight tensors in contiguous memory regions, \ie dense layers remain dense in all lower-capacity subnetworks. Compared to standard pruning methods, which encode structural sparsity patterns using binary masks and apply uniform pruning ratios across all layers, \reds enforces that all retained neurons or filters form a single contiguous block in each weight tensor. Each \reds subnetwork can be specified by simple (offset, length) parameters without storing any masks. Moreover, \reds computes layer-specific pruning ratios and these are determined based on each layer’s contribution to the model’s representational capacity and the cost associated with application-specific computational constraints. \reds enables precise control over inference costs while preserving representational performance. In contrast to \citet{shen2022structural}, \reds knapsack formulation models the dependency between layers by iteratively chaining constraints to enforce each subnetwork's weight tensor dimensions to be functionally correct (see \figref{fig:REDS_Convolution_splits.png}). Furthermore, \reds knapsack formulation can be extended to constrain each subnetwork's peak memory usage, \ie activation memory usage, defined as the major memory bottleneck for enabling neural networks on the edge computing devices
~\citep{lin2023tiny}. Our theoretical findings in the Appendix suggest that the knapsack method applied to the smallest subnetwork first and letting the \reds structure iteratively grow, \ie the \emph{bottom-up approach}, is more effective than starting from the largest subnetwork and pruning it down to the smallest one, \ie the \emph{top-down approach}. We empirically show this in \secref{sec:reds:performance}.

\reds code is publicly available.\footnote{Link: \url{https://github.com/FraCorti/REDS_TOMC}}
Our contributions are:
\begin{itemize}
	\item We present a novel hardware-aware method to convert a model into the \reds structure, which can efficiently adapt to dynamically changing resource constraints.   
    \item We formulate the optimization problem as an iterative knapsack problem~\citep{DELLACROCE201926}, define a novel generalized knapsack framework for depth-wise convolutions and vision transformers, present theoretical analysis and provide empirical evidence of the solution effectiveness, especially in the low-data regime (\secref{sec:reds} and \secref{sec:reds:performance}).
    \item \reds uses the permutation invariance of neurons to facilitate hardware-specific optimizations (\secref{sec:reds}). In particular, for resource-constrained devices that use data caches, \reds provides a compile-time optimization to ensure all subnetwork weights reside in contiguous memory (\secref{sec:cache}). 
     
\end{itemize}

In the next section we explain \reds on a fully-connected neural network (\dnn).

\section{\reds Applied to a Simple Network}
\label{sec:sample}

\begin{figure*}
    \centering
    \includegraphics[width=\textwidth]{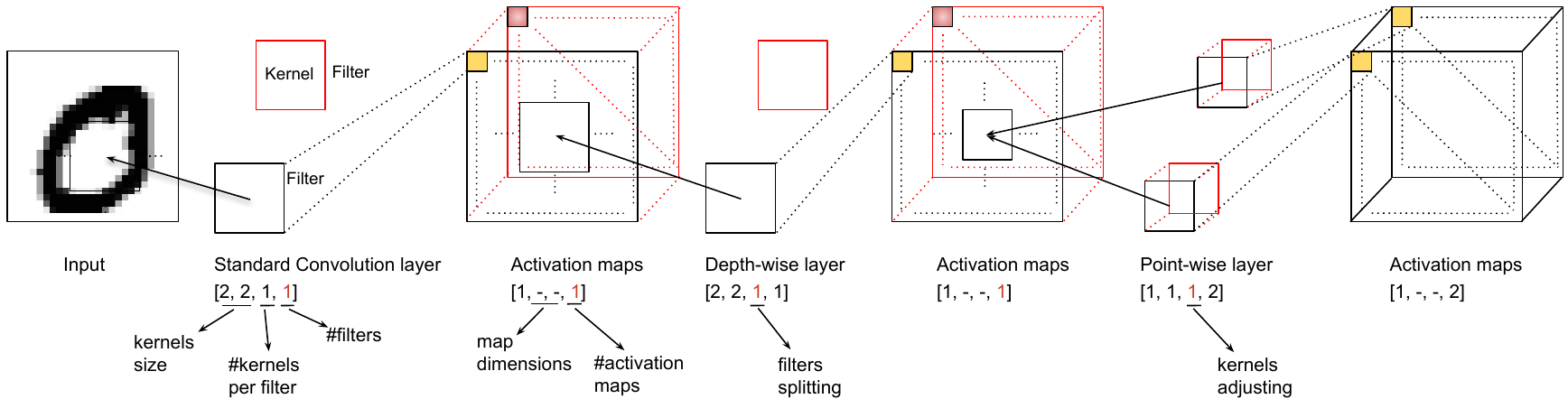}
    \caption{Filter removal in a \dscnn model with one standard convolutional layer and one depth-wise and point-wise convolutional block with layers width equal to two.  
    The dimensions of the sliced filters, the sequence of the reorganized kernels, and the weight tensors chaining constraints are highlighted in red.}  
    \label{fig:REDS_Convolution_splits.png}
\end{figure*}

We first illustrate advantages of \reds on a fully-connected network depicted in \figref{fig:REDS_Dense_splits}. The three dense weight matrices $W_1^{3\times 4}$, $W_2^{4\times 4}$ and $W_3^{4\times 2}$ connect neurons in consecutive layers that are stored in memory as unfolded one-dimensional arrays. We assume that the row-major format is used to store and access model weights in memory, which is the standard choice on most hardware architectures. Given a trained and optimized model, weight matrices are mapped to continuous memory regions, allowing for a straightforward use of vector instructions to speed up on-device inference.

\reds organizes a model into a set of \emph{nested submodels}, \ie the active weights of a child subnetwork are fully contained in its parent subnetwork. To enable this structure, we slice each parent subnetwork into an \emph{active} and an \emph{inactive} part, where the active part shapes the child subnetwork. \reds uses structured sparsity, \ie model slicing occurs at the level of individual neurons and convolutional filters. Slicing off two neurons in the first hidden layer in \figref{fig:REDS_Dense_splits} leads to removing two columns and two rows in the weight matrices $W_1$ and $W_2$, respectively.

First, although any neuron can be removed from a layer, we re-order neurons to have a group of active neurons followed by inactive neurons due to the \emph{permutation invariance} phenomenon of neural networks~\citep{entezari2021role}. A permutation does not change the function of the network, but allows optimizing the memory layout to keep subnetwork weights in contiguous memory. The above observations make the subnetwork layout to be stored efficiently on-device. In fact, it is sufficient to store one integer value denoting the \emph{slicing point} for each layer in each subnetwork, corresponding to the number of active computational units. This is possible since active and inactive units build contiguous groups in memory. Activating a particular subnetwork means changing the size of the dimension of the weight tensor in each layer. An additional bit indicates optimization for caches, \ie that a flipped operation order is applied.

Second, \reds achieves \emph{minimal adaptation overhead}: only the width of the layers has to be updated to switch to a different model. Switching from one \reds submodel to another requires adjusting only the width sizes of the layers. Due to the local scope of a slice, the modification affects only the incoming and outgoing connections; inactive computational units do not participate in inference. Examples of slicing dense and convolutional networks are shown in \figref{fig:REDS_Dense_splits} and \figref{fig:REDS_Convolution_splits.png}. The computational cost of running inference using a subnetwork is not affected by the presence of other \reds networks, in sharp contrast to the approaches that use binary masks to select active neurons or channels \cite{qu2022dress, mishra2021accelerating}.

In the end, pruning a neuron in one layer may remove more multiply–accumulate operations (MACs) than in another layer. In our example in \figref{fig:REDS_Dense_splits}, removing one neuron in the first hidden layer yields more MAC reduction than removing a neuron in the second hidden layer due to a different number of connections. Moreover, the contribution of these neurons to model accuracy may be different. Related research suggests several importance measures for individual weights, neurons, and convolutional filters~\citep{shen2022structural,molchanov2019importance,Hoefler2021sparsity,frantar2022spdy}. These can be used to optimize a network for performance while keeping essential elements. Our problem is different: we build a nested structure, which requires iterating over each subnetwork. Previous approaches~\citet{fang2018nestdnn, qu2022dress} start from the largest network or implement a top-down pruning and a bottom-up freeze-and-regrow algorithm. We formulate the optimization problem for a parent-child subnetwork pair as a variant of the knapsack problem~\citep{shen2022structural}. To extend the solution to multiple nested subnetworks, we generalize the approach to an iterative knapsack problem~\citep{DELLACROCE201926}. 

In the following sections, we focus on specific contributions and design choices for \reds on the way to a fully-functioning framework. \secref{sec:reds} discusses a pipeline to convert a model to \reds and presents a hardware-aware iterative knapsack method to choose the model slicing points. \secref{sec:cache} revisits the memory layout optimization for data cache and provides an empirical evaluation of the achieved gain on embedded and mobile hardware platforms.

\section{Resource-Efficient Deep Subnets}
\label{sec:reds}

Before detailing the \reds construction pipeline, we first argue that layer slicing by skipping a \emph{contiguous} sequence of computational units, whether neurons or convolutional filters, from any layer except the final classifier, preserves the representational capacity of the resulting \reds subnetworks. The pruned computational units can be reordered to form a contiguous sequence without changing the network function. We then describe \reds construction pipeline. At the core is the novel method to decide on the choice of the subnetwork structure of \reds based on the knapsack problem. Finally, we discuss how to fine-tune the submodels in \reds to recover lost accuracy due to the applied structural pruning of computational units. 

\vspace{-0.2cm}
\subsection{Permutation invariance}
\label{sec:permutation_invariance}
The optimization landscape of neural networks contains an abundance of minima and saddle points as a result of numerous symmetries. One important class of symmetries is formed by permutations, \ie the neurons or filters in the same layer can be randomly permuted and, if the incoming and outgoing connections are adjusted accordingly, form structurally different, yet functionally identical networks. An example is given in~\secref{sec:sample}. A layer with $n$ neurons has $n!$ permutations. A deep network with $l$ such layers has $\prod_l n!$ functionally identical solutions~\citep{entezari2021role}.

We leverage permutation invariance of neural networks to permute the units of a layer in descending order based on their importance scores. This operation does not change the subnetwork function, but ensures that the nested subnetworks keep the most important units of each layer as part of their architecture. To preserve the correct feature extraction process, the permutation operation is performed layerwise and the structural elements to permute depend on the layer type. In a convolutional layer, for example, the weights are stored as a four-dimensional tensor (see Fig.~\ref{fig:REDS_Convolution_splits.png}). When the convolutional filters are permuted, the kernels in the subsequent layer are similarly reordered to maintain consistent input channel and kernel sequence during the forward pass. This permutation process involves manipulating the original weight tensors, initially reshaping the layer's tensor into a set of three-dimensional tensors representing the convolutional filters. Each filter tensor is then reshaped into a set of two-dimensional tensors, each corresponding to one of the filter's kernels.

\subsection{Cost measures}
How do we identify key neurons or convolutional filters essential for constructing a high-precision subnetwork? Several studies use importance scores to quantify the significance of network weights, neurons, and channels.  
NestDNN~\citep{fang2018nestdnn} ranks each convolutional filter by calculating the L2 distance between feature maps of the same class and those of different classes. This method uses all the training data, which is often unavailable due to resource constraints when the model is deployed on edge devices. Magnitude-based pruning methods, that use the magnitude of the weights to estimate their importance~\citep{han2015learning, zhu2017prune, cai2020onceforall}, provide a computationally efficient way to rank the model's computational units. However, several studies~\citep{molchanov2019importance, mozer1988skeletonization} have reported that the magnitude does not necessarily reflect the importance of a unit in terms of its contribution to the loss reduction.

In \reds, we compute for each of the encoder's weight $i$ the importance as $I_{i} = | g_{i} \gamma_{i}|$, where $g_{i}$ is the sum of the accumulated gradients computed from backpropagation~\citep{rumelhart1986learning} and $\gamma_{i}$ is the scalar value of weight $i$~\citep{shen2022structural}. Weights with larger accumulated gradients $g_{i}$ contribute more to reducing the cross-entropy loss function, making them critical for model's accuracy preservation~\citep{molchanov2019importance}. For each computational unit $c$, such as a convolutional filter or a fully-connected neuron, let $\mathcal{W}_c$ be the set of its weights. Then its importance score, denoted as $I_c$, is computed as the grouped sum of the importance scores over all these weights:
\begin{equation}
\label{eq:computational_unit_score}
    I_{c} = \sum_{i \in \mathcal{W}_c} I_{i} = \sum_{i \in \mathcal{W}_c} | g_{i} \gamma_{i}|.
\end{equation}
The computational unit \impscore approximates the squared difference of the loss when a unit is removed; thus it quantifies the contribution of each unit to the model's accuracy~\citep{molchanov2019importance}. Each computational unit is characterized by an importance score, denoted as $I_{c}$, and computational costs defined in terms of model latency and memory usage. We compute inference latency and memory usage predictors for each unit on an edge device. Inference latency is defined as the number of multiply–accumulation operations (MACs)~\citep{liberis2021munas}, while memory usage is calculated as the size in bytes of the output tensor. 

\reds adopts the gradient-based importance score formulation introduced by~\citep{molchanov2019importance} reported in~\equref{eq:computational_unit_score}. Related unit scoring methods such as SNIP~\citep{lee2018snip}, which computes filter importance scores using a similar gradient-based metric, and OSTR~\citep{yang2024ostr}, which derives operation-strength scores from the architecture-parameter gradients during neural architecture search, have also been proposed. However, a detailed comparison with these methods is beyond the scope of this work and is left for future investigation.

\subsection{Bottom-up (\bottomup) and top-down (\topdown) heuristics}
\reds examines two heuristics for solving the iterative knapsack problem named bottom-up (\bottomup) and top-down (\topdown). The former iteratively calculates the subnetwork architectures by considering the tightest constraints for the smallest subnetwork first. Once a solution is found by the knapsack solver, these units are frozen, \ie they are now part of all nested subnetworks, and the second smallest subnetwork is being computed by the solver. The latter top-down method determines solutions by considering the weakest constraints first and then iteratively searching the architectures for increasingly smaller subnetworks. Related work~\citet{cai2020onceforall} uses a variant of the top-down approach, \ie progressive shrinking. Our theoretical analysis of the iterative knapsack problem (based on the classical 0-1 knapsack problem) in the Appendix shows that the bottom-up approach promises a better worst-case performance. In particular, we prove that the solution found by the two-stage bottom-up iterative knapsack heuristic is not worse than $\frac{2}{3} \cdot Opt$, where $Opt$ is the optimal solution of the knapsack with full capacity, and that this bound is tight. We then show that the tight bound for the top-down iterative knapsack is $\frac{1}{2} \cdot Opt$, where $Opt$ in this case is the optimal solution of the knapsack with reduced capacity. Since our generalized problem suited for depthwise
separable convolutional architectures has the classical 0-1 knapsack as its core problem, we believe that a similar result is valid for this case as well.

\subsection{Iterative knapsack problem}
\label{sec:iterative_knapsack}
This section describes the novel method to design \reds subnetwork structure by formulating and solving a variant of an iterative knapsack problem. The theoretical analysis is moved to the Appendix \ref{appendix:iterative_knapsack} for presentation clarity.

\subsubsection{Problem formulation} 
Given a pre-trained model, \reds identifies how to best slice the weight tensors by formulating the problem as an iterative knapsack problem with $k$ stages: the items included in a knapsack with capacity $c$ have to be included in all later stages, \ie knapsacks with larger capacities~\citep{DELLACROCE201926, Faenza23}. The items correspond to all the computational units that compose the model encoder architecture, \ie filters and kernels.
Given $C_{MACs}$ as the maximum number of MACs, for a subnetwork $s$, a single stage knapsack problem is formulated as follows:
\begin{equation}
\begin{aligned}
& \max && \sum_{l=1}^L \sum_{i=1}^{u_l} x_{il} \cdot I_{il}  && \\
&\text{s.t.}&& \sum_{l=1}^L \sum_{i=1}^{u_l} x_{il} \cdot MACs_{il} \leq C_{MACs}\,, \\
&&& x_{il} \in \{0, 1\}, \quad \forall\, l \in \{1,\dots,L\}, \;\; i \in \{1,\dots,u_l\}.
\end{aligned}
\label{eq:knapsack_first}
\end{equation}
where $x_{il}$ is a binary decision variable taking value one if an item $i$ from layer $l$ is selected and zero otherwise; $I_{il}$ is the importance score of item $i$ from layer $l$; $MACs_{il}$ is the number of MACs of item $i$ from layer $l$; $L$ is the number of encoder layers in the model and $u_l$ is the number of items in layer $l$.  
As an extension of this model we added the constraint to each knapsack problem that one unit per layer must be taken. 
This guarantees that the subnetworks extraction will not degenerate into a trivial solution, \ie pruning whole layers.
\def\dw{d}
\def\pf{f}
\def\pft{g}

\subsubsection{Knapsack for depth-wise convolutions}
\label{appendix:knapsack_variant_two}

We define a novel generalized knapsack framework for depth-wise convolutions~\citep{howard2017mobilenets}. 
Given a $m \times n$ input matrix $A$, the first layer $L_0$ is a standard convolutional layer (\cf Fig.~\ref{fig:REDS_Convolution_splits.png}) applied to $A$ that consists of $N_0$ filters and hence produces $N_0$ output channels. 
It is followed by depth-wise convolutional blocks $B_1, \ldots, B_d$, where each block $i$ is built by a depth-wise layer $B_i^D$ and a point-wise layer $B_i^P$ for $i=1,\ldots, d$.
A depth-wise layer applies one filter on each of the $N_{i-1}$ input channels. 
The point-wise layer $B_i^P$ then applies $N_i$ filters, where the kernel number for each filter has to be equal to the number of filters of the previous layer. 
If we want to reduce the size of the network, we can decide on how many filters we use at $L_0$ and at each $B_i^P$. 
E.g., if we choose $k$ filters at $L_0$, the layer $B_1^D$ will have $k$ filters and one filter of $B_1^P$ will have $k$ as kernels number. 
Integer programming allows structural pruning of a neural network, \ie choosing an optimal number of filters of $L_0$ and of $B_i^P$ to maximize performance of the network obeying a constraint on the number of MACs.

Integer decision variables $x_0$ are used for the number of filters at $L_0$, and $x_i$ for the number of filters at each $B_i^P$. 
Other decision variables control whether a unit is used by a subnetwork or not. 
For $L_0$ we introduce $N_0$ binary variables $y_1, \ldots, y_{N_0}$, and for every block $B_i$, $i=1,\ldots, d$, binary variables $\pf_k^i$ and $\pft_{kt}^i$, $k\in \{1,\ldots, N_{i}\}$ and $t\in \{1,\ldots, N_{i-1}\}$, $\pf_k^i$ to indicate whether a filter $k$ is used in $B_i^P$, and $\pft_{kt}^i$ if filter $k$ with kernel $t$ of $B_i^P$ is used. 
For $B_i^D$ binary variables $\dw_t^i$ decide if a depth-wise filter $t \in \{1, \ldots, N_{i-1}\}$ is used. $P^1$ is the importance score of a standard convolution filter $i$ in the first layer and $W_1$ its number of MACs. 
For each depth-wise point-wise block $i$, $P^i_{t}$ is the importance score for the depth-wise filter $t$, $W_2^i$ its number of MACs, $P^i_{kt}$ is the importance score of the corresponding kernel $k$ of the point-wise filter $t$ in the subsequent layer $N_i$ and $W_3^i$ its number of MACs.

The knapsack formulation for a single stage problem of depth-wise convolutions is described by the following ILP-model, where all variables are binary. 
\begin{itemize}
    \item[\eqref{eq:(1)}] Maximization of the total importance score of the chosen architecture. 
    \item[\eqref{eq:(2)}] Satisfaction of constraint $C$ (\ie MACs).
    \item[\eqref{eq:(3-4)}] The number of filters chosen in the first convolution layer is represented by $x_0$, and the number of filters chosen in the depth-wise layer $i$ must match the number of filters picked in the previous layer $i-1$. 
    \item[\eqref{eq:(5-6)}] The number of the point-wise filters chosen in layer $i$ is set to $x_i$, and the point-wise filters are chosen starting at the first point-wise filter to be contiguous in memory.
    \item[\eqref{eq:(7-8)}] If kernel $t$ of filter $k$ is chosen then the whole filter $k$ is chosen.
    A point-wise filter $k$ is chosen only if one of its kernels is chosen.
    \item[\eqref{eq:(9)}] The number of kernels in the filter $k$ of point-wise layer $i$ must be less or equal to the number of filters taken in the previous depth-wise layer.
    \item[\eqref{eq:(10)}] If filter $k$ in layer $i$ is chosen then constraints \eqref{eq:(9)} and \eqref{eq:(10)} ensure that the number of kernels $t$ of filter $k$ in layer $i$ equals the number of point-wise filters in the previous block. If filter $k$ in layer $i$ is not chosen, constraints  \eqref{eq:(9)} and \eqref{eq:(10)} imply that all kernels $t$ of filter $k$ at layer $i$ are zero ($N_{i-1}$ is an upper bound on $x_{i-1}$).
\end{itemize}

{\small 
\begin{eqnarray}
& \max & \sum_{i=1}^{N_0} y_i \cdot P^1 + \sum_{i=1}^{d} \sum_{t=1}^{N_{i-1}} \left(\dw_t^i \cdot P^i_{t} + \sum_{k=1}^{N_{i}} \pft_{kt}^i \cdot P^i_{kt} \right)  \label{eq:(1)} \\  
& \text{s.t.} & \sum_{i=1}^{N_0} y_i \!\cdot\! W_1 + \!\!\sum_{i=1}^{d} \!\sum_{t=1}^{N_{i-1}} \!\!\left(\dw_t^i \!\cdot\! W_2^i + \!\!\sum_{k=1}^{N_{i}}  \pft_{kt}^i \!\cdot\! W_3^i \!\right) \!\leq C \label{eq:(2)}\\
&& \sum_{i=1}^{N_0} y_i = x_0 \; \;\;\;\;\text{and}\;\;\;\;
\sum_{t=1}^{N_{i-1}} \dw_t^i = x_{i-1} \enspace \forall i \label{eq:(3-4)} \\
&& \sum_{k=1}^{N_{i}} \pf_k^i = x_i \enspace \forall i \;\;\;\;\;\text{and}\;\;\;\;
 \pf_k^i \geq \pf_{k+1}^i \enspace \forall i,k \label{eq:(5-6)} \\
&& \pft_{kt}^i \leq \pf_k^i\enspace \forall i,k,t \;\;\;\;\;\text{and}\;\;\;\;
\pf_k^i \leq \sum_{t=1}^{N_{i-1}} \pft_{kt}^i \enspace \forall i,k \label{eq:(7-8)} \\
&& \sum_{t=1}^{N_{i-1}} \pft_{kt}^i \leq x_{i-1} \enspace \forall i,k \label{eq:(9)} \\ 
&& \sum_{t=1}^{N_{i-1}} \pft_{kt}^i \geq x_{i-1} - (1-\pf_k^i) \cdot N_{i-1} \enspace \forall i,k \label{eq:(10)}
\end{eqnarray}
}

\subsubsection{Knapsack for Vision Transformers}
\label{sec:knapsack_variant_vision_transformers}
We define a generalized knapsack framework for Vision Transformer~\citep{dosovitskiy2020image}. Given a $c \times h \times w$ input matrix $A$, the first block $P_0$ is a patch embedding block using a convolutional layer to linearly project $A$ into a sequence of patches. The patch embedding convolutional layer consists of convolutional filters whose embedding dimension corresponds to the $N_0$ attention heads used in the subsequent \vit layers. The patch embedding block is followed by $B_1, \ldots, B_d$ blocks, where each block $t$ is composed by a Multi-Head Self-Attention (\msa) layer $B_t^{MSA}$ and two  multi layer perceptron layers (\ie linear layers) $B_t^{\mathrm{mlp1}}$ and $B_t^{\mathrm{mlp2}}$ for $t=1,\ldots, d$. Each \msa layer is composed of a $MSA_{qkv}$ linear layer consisting of the concatenation of the query, key and value matrices with their corresponding attention heads. It is followed by a linear projection layer $MSA_{proj}$. Residual connections enforce that the input and output latent patch vectors of each \msa and \mlp layer have the same dimension. Structural pruning of a \vit model via integer programming is formulated as choosing the optimal number of attention heads' convolutional filters $N_0$, that reduce each patch dimension of sixty-four for each head's convolutional filters pruned~\citep{haberer2024hydravit}, and the number of neurons of each $B_i^{\mathrm{mlp1}}$ to maximize the performance of the network obeying a constraint (\ie number of MACs).  

Integer decision variables $x_0$ are used for the number of filters at $P_{0}$ and variables $n_{\mathrm{mlp1_t}}$ for the number of neurons at each $B_t^{\mathrm{mlp1}}$. For $P_{0}$ we introduce $N_{0}$ binary variables $y_1, \ldots, y_{N_0}$, for every block layer   $B_t^{\mathrm{mlp1}}$, $t=1,\ldots, d$, binary variables $f_{ik}^t$ and for every block layer $B_t^{\mathrm{mlp2}}$ binary variables $g_{ki}^t$ where $i\in \{1,\ldots, N_0\}$ and $k\in \{1,\ldots, N_{i}\}$, $f_{ik}^t$ to indicate whether weights $k$ in the neuron $i$ of $B_t^{\mathrm{mlp1}}$ are used, and $g_{ki}^t$ to indicate whether weights $i$ in the neuron $k$ of $B_t^{\mathrm{mlp2}}$ are used.  

$P_{pat_{i}}$ is the importance score of the head $i$ convolutional filters in the patch embedding block and $W_1$ its number of MACs. For each block $t$, $P_{Q_{i}}^t$, $P_{V_{i}}^t$ and $P_{K_{i}}^t$ are the importance scores of the head $i$ in the qkv linear layer, $P_{ik}^{\mathrm{mlp1}_{t}}$ is the importance score of neuron $i$ weight $k$ in the $\mathrm{mlp1}_t$ layer and $P_{ki}^{\mathrm{mlp2}_t}$ is the importance score of neuron $k$ weight $i$ in the $\mathrm{mlp2}_t$ layer, $W_{ik}^{\mathrm{mlp1}_{t}}$ and $W_{ki}^{\mathrm{mlp2}_{t}}$ are their number of MACs. 

The knapsack formulation for a single stage problem of \vit is described by the following ILP-model, where all variables are binary. 
\begin{itemize}
    \item[\eqref{eq:(11)}] Maximization of the total importance score of the chosen \vit architecture. 
    \item[\eqref{eq:(12)}] Satisfaction of constraint $C$ (\ie MACs).
    \item[\eqref{eq:(13)}] The number of convolutional filters for each head in the patch embedding layer is represented by $x_0$ and the number of neurons in the multi-layer perceptron layer $\mathrm{mlp1}_t$ of block $t$ is represented by $n_{\mathrm{mlp1}_t}$. 
    \item [\eqref{eq:(13a)}] The number of columns of the first  neuron of $\mathrm{mlp1}_t$ (\ie $\sum_{k=1}^{N_o} f_{1k}^t$) is equal to number of chosen convolutional filters $x_0$.
    \item[\eqref{eq:(14)}] The number of columns of the first  neuron of $\mathrm{mlp2}_t$ (\ie $\sum_{k=1}^{N_i} g_{1k}^t$) is equal to the number of chosen $n_{\mathrm{mlp1}_t}$ neurons in $\mathrm{mlp1}_t$.
    \item[\eqref{eq:(15)}] If a neuron $f_i$ is picked, then its number of input weight connections is equal to the number of chosen heads convolutional filters  $x_0$. 
    \item[\eqref{eq:(16)}] The number of neurons chosen in $\mathrm{mlp2}_t$ is equal to the number of heads convolutional filters chosen $x_0$, and the sum of each neuron input weights is equal to the number of neurons chosen in the previous $\mathrm{mlp1}_t$ layer.
    \item[\eqref{eq:(17)}] The number of heads convolutional filters $N_0$ are chosen to be contiguous in memory.
    \item[\eqref{eq:(18)}] If head convolutional filters $y_i$ is not chosen then all the corresponding columns in $\mathrm{mlp1}_t$ and rows in $\mathrm{mlp2}_t$ are equal to zero. 
    \item[\eqref{eq:(19)}] The number of neurons in $\mathrm{mlp1}_t$ and $\mathrm{mlp2}_t$ are chosen to be contiguous in memory. 
    \item[\eqref{eq:(20)}] The number of each neuron's weight in $\mathrm{mlp1}_t$ and $\mathrm{mlp2}_t$  are chosen to be contiguous in memory. 
\end{itemize}
{\small 
\setlength{\arraycolsep}{2pt}
\begin{eqnarray}
& \max & 
\sum_{i=1}^{N_o}\Bigl(
  y_i\cdot P_{\mathrm{pat}_{i}}
+ \sum_{t=1}^{d} y_i\cdot \Bigl( P_{Q_{i}}^t + P_{V_{i}} + P_{K_{i}}^t + P_{\mathrm{proj}_{i}}^t 
\Bigr) \Bigr)
\nonumber\\
&&\quad
+ \sum_{t=1}^d \sum_{i=1}^{N_o} \sum_{k=1}^{N_i}
   f_{ik}^t\;\cdot\;P_{ik}^{\mathrm{mlp1}_{t}}
\nonumber\\
&&\quad
+ \sum_{t=1}^d \sum_{i=1}^{N_o} \sum_{k=1}^{N_i}
   g_{ki}^t\;\cdot\;P_{ki}^{\mathrm{mlp2}_{t}}
\label{eq:(11)}\\
&\text{s.t.}&
\sum_{i=1}^{N_o}
  \Bigl( y_i\cdot W_{\mathrm{pat}_{i}} + \sum_{t=1}^{d} y_i\cdot \Bigl(W_{Q_{i}}^t + W_{V_{i}}^t + W_{K_{i}}^t + W_{\mathrm{proj}_{i}}^t \Bigr) \Bigr) 
\nonumber\\
&&\quad
+ \sum_{t=1}^d \sum_{i=1}^{N_o} \sum_{k=1}^{N_i}
   f_{ik}^t\;\cdot\;W_{ik}^{\mathrm{mlp1}_{t}}
\nonumber\\
&&\quad
+ \sum_{t=1}^d \sum_{i=1}^{N_o} \sum_{k=1}^{N_i}
   g_{ki}^t\;\cdot\;W_{ki}^{\mathrm{mlp2}_{t}}
\;\leq\;C
\label{eq:(12)}
\end{eqnarray} 
\setlength{\arraycolsep}{2pt}
\begin{eqnarray}
    &&
\sum_{i=1}^{N_o} y_i = x_0 \;\;\;\;\text{and}\;\;\;\; \sum_{i=1}^{N_i} f_{i1}^t = n_{\mathrm{mlp1}_t} \enspace
\forall t 
\label{eq:(13)}\\
&&
\sum_{k=1}^{N_o} f_{1k}^t = x_0 \enspace \forall t 
\label{eq:(13a)}\\
&&
\sum_{k=1}^{N_i} g_{1k}^t = n_{\mathrm{mlp1}_t} \enspace \forall t
\label{eq:(14)}\\
&&
\sum_{i=1}^{N_o} f_{ik}^t  \geq x_0 - N_o \cdot (1 - f_{1k}^t) \enspace \forall k,t 
\label{eq:(15)}\\
&&
\sum_{i=1}^{N_o} g_{ik}^t \geq x_0 - N_o \cdot (1 - g_{1k}^t) \enspace \forall k,t 
\label{eq:(16)}\\
&&
y_{i} \ge y_{i+1} \enspace \forall i
\label{eq:(17)}\\
&&
f_{ik}^t \leq y_{i}\;\;\;\;\;\;\;\;\;\;\text{and}\;\;\;\; g_{ik}^t \leq y_i \;\enspace\quad\quad\forall i,k,t
\label{eq:(18)}\\
&&
f_{ki}^t \ge f_{i,k+1}^t
\;\;\;\;\text{and}\;\;\;\;
g_{ik}^t \ge g_{i,k+1}^t
\quad\forall i,k,t
\label{eq:(19)}\\
&&
f_{ik}^t \ge f_{i+1,k}^t
\;\;\;\;\text{and}\;\;\;\;
g_{ik}^t \ge g_{i+1,k}^t
\;\quad\forall i,k,t
\label{eq:(20)}
\end{eqnarray}
}

\subsubsection{Peak Memory Usage Constraint}
Formulating structural pruning of a neural network as an iterative knapsack problem allows to add constraints to each subnetwork’s architecture by assigning different computational costs to each item in the integer program. We show how to constrain peak memory usage by limiting the summed size of the output tensors for each layer~\citep{liberis2023differentiable}. Given the previously defined single-stage knapsack formulation \eqref{eq:knapsack_first}, $OutMem_{il}$ as the size of the output tensor of item $i$ from layer $l$, and $C_{PeakMem}$ as the maximum activation map size, the peak memory usage constraint is defined as follows:  

\begin{equation}\label{eq:peakmem}
    \begin{aligned}
        \sum_{i=1}^{u_l} x_{il} \cdot OutMem_{il}  \leq C_{PeakMem} \enspace \forall  l\\ 
    \end{aligned}
\end{equation}

\subsubsection{Knapsack for Language and Multimodal Models}
\reds knapsack formulation for \vit~\secref{sec:knapsack_variant_vision_transformers} can be extended to language models (\ie \llms) and multimodal models. In fact, these models often comprise transformer blocks composed by multi-head self-attention and multilayer perceptron (\ie \mlp) layers. For example, Llama2~\citep{touvron2023llama} uses transformer blocks with a higher number of attention heads (\ie 32) and larger embedding dimensions (\ie 4096) than the \vit-Base configuration~\citep{dosovitskiy2020image}. Moreover in Llama2, for each transformer block $B^t$, the weights of each attention head (\ie $W_{Q_i}^t + W_{K_i}^t + W_{V_i}^t + W_{\mathrm{proj}_i}^t$), the neurons of the $\mathrm{mlp1}_t$ layer, and the neurons of the $\mathrm{mlp2}_t$ layer can be treated as items in the \reds iterative knapsack formulation. For multimodal models that comprise vision and text encoders, \eg \vit models~\citep{li2024scaling}, the \reds knapsack formulation can be modelled by considering each encoder architecture computational unit as part of the knapsack formulation with their respective constraints, computational costs and scores. Solving the \reds iterative knapsack formulation for these models yields a set of subnetworks that satisfy the specified computational constraints. 

Extending \reds to \llm such as Llama2 may lead to combinatorial growth in the number of candidate subnetwork architectures. For example, Llama2-7B comprises 32 transformer blocks, each block $t$ contains 32 attention heads, 11008 $\mathrm{mlp1}_t$ and 4096 $\mathrm{mlp2}_t$ neurons, which can render the knapsack formulation computationally infeasible for obtaining optimal solutions. To ensure tractability, one can group heads or neurons into coarser blocks to reduce the search space. Alternatively,
one can use a different strategy for solving the knapsack variant under consideration. Note that in combinatorial optimization, similar problems are frequently solved by tailor-made heuristics or meta-heuristics. In this case, the solution comes with no proven optimality guarantee, but is usually of very high quality, see \eg \citet{CACCHIANI2022105692}, \citet{wilbaut2008survey}, \citet{cacchiani2022knapsack}, \citet{laabadi20180} for knapsack variants and corresponding heuristics and meta-heuristics. Given a subset-selection model for \llm architectures, such a heuristic or meta-heuristic could be designed to identify a promising subnetwork. Additionally, the solution could then be compared to the optimal solution obtained by the \reds knapsack formulation on a smaller, time-feasible (with respect to an ILP) problem instance. A detailed investigation of \reds knapsack formulation extensions for \llm and multimodal models is left for future work.

\subsubsection{Knapsack for Semantic Segmentation and  Object Detection Models}
\reds knapsack formulation is agnostic to the architecture of the input model, provided that the model and its loss function are differentiable, thus allowing gradient computation, required when employing gradient-based importance scores for the units. Given a pre-trained segmentation or object-detection model, \reds computes unit-wise importance scores using the respective task-specific loss (\eg pixel-wise cross-entropy for segmentation or the combined classification and localization loss for detection) and then solves the iterative knapsack problem under specified computational or memory budget constraints given as input. The selected subnetworks can subsequently be fine-tuned, using the same task-specific loss, to ensure high predictive performance on the target task.

\subsection{\reds fine-tuning}
Fine-tuning models after pruning is a common practice to recover model accuracy. The \reds subnetworks found by the heuristics algorithm are fine-tuned to recover their full accuracy. For the \dscnn, \cnn and \dnn architectures we first simultaneously fine-tune the encoder's weights followed by the batch normalization layers~\citep{qu2022dress}. For the \vit architecture we employed a stochastic training approach following~\citep{haberer2024hydravit} to finetune each subnetwork. For each subnetwork we slice each layer to construct a subnetwork architecture by storing only the slicing point corresponding to the number of active computational units. 
During fine-tuning, the weights of each subnetwork $i$ are reused by all lower-capacity subnetworks $\{i+k\}_{k=1}^{N}$ by dynamically creating a tensor slice for each layer, \ie a tensor view object that points to the original weight tensor. For the \dscnn, \cnn and \dnn architectures we balance the contribution of the loss of each individual model with parameters $\{\pi_i\}_{i=1}^{N}$ equal to the percentage of weights used in the encoder part of the model by the subnetwork $i$~\citep{qu2022dress}. 

\begin{figure*}[t!]
\includegraphics[width=0.24\textwidth]{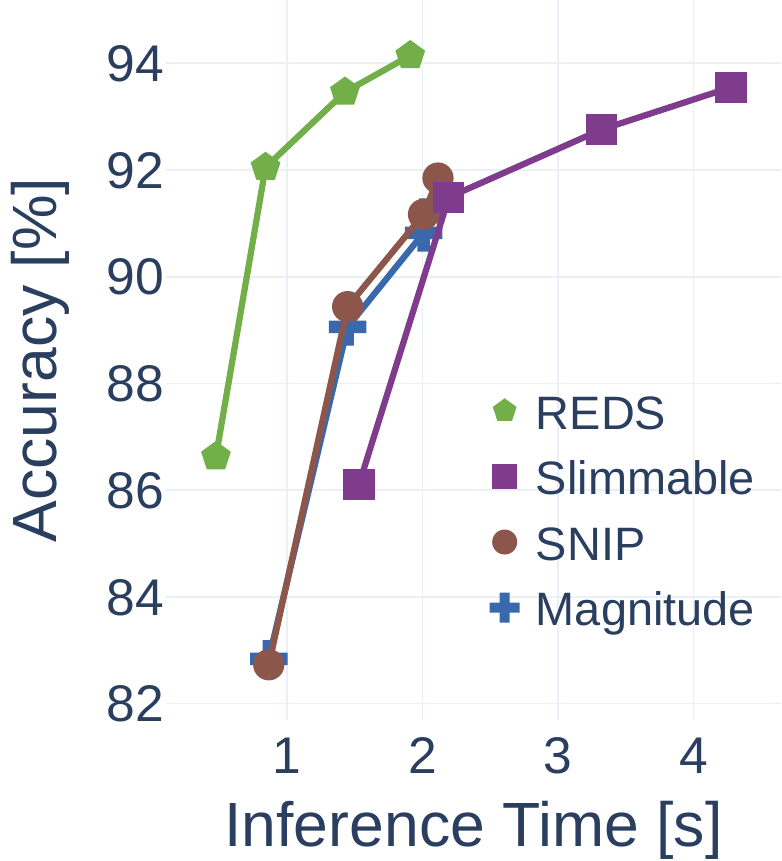}
\includegraphics[width=0.24\textwidth]{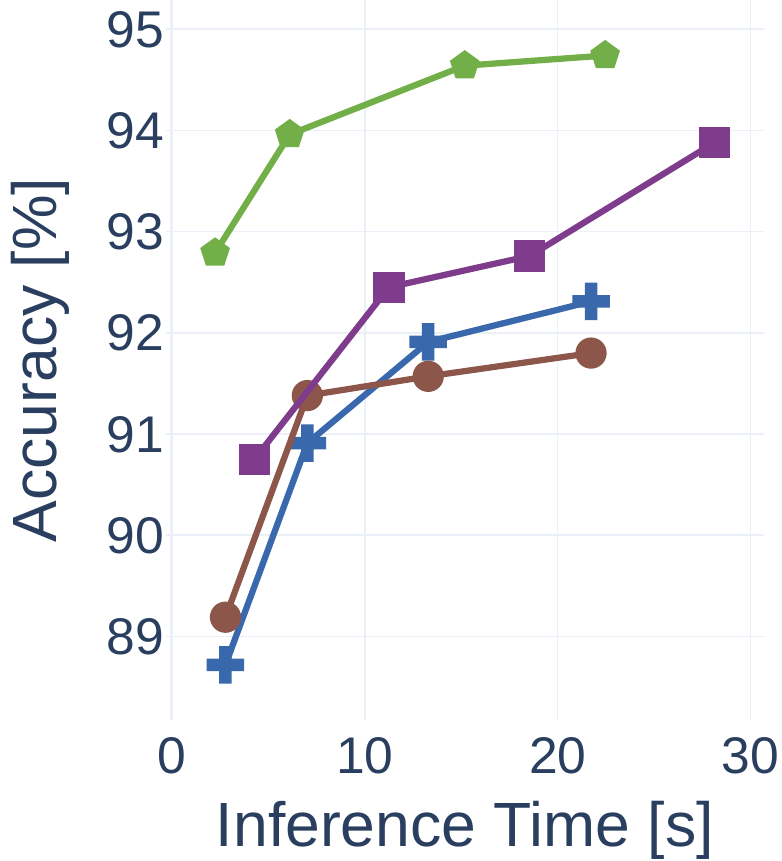}
\includegraphics[width=0.24\textwidth]{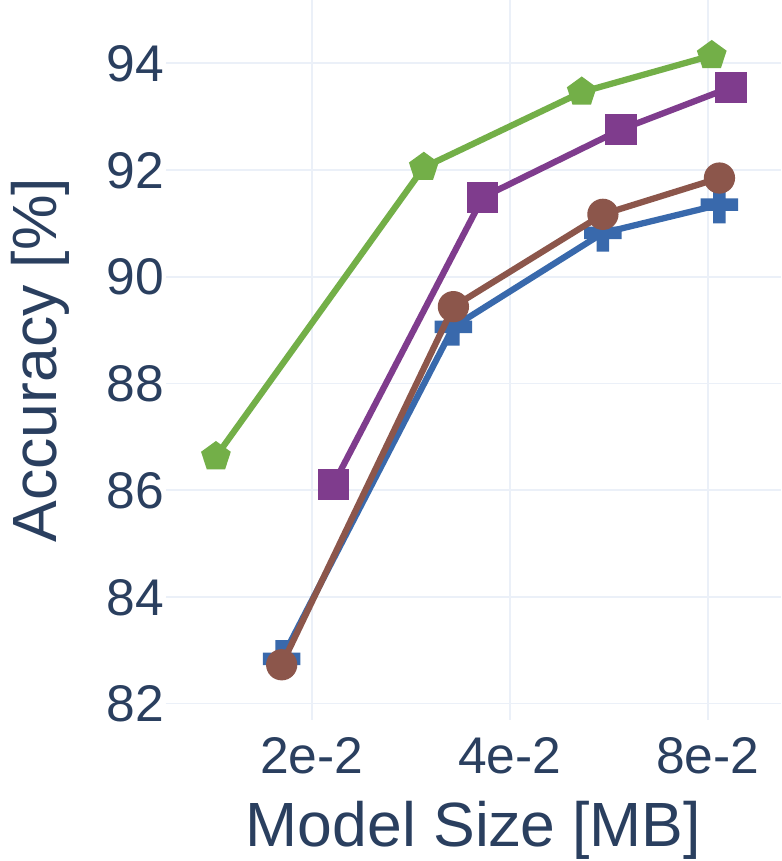}
\includegraphics[width=0.24\textwidth]{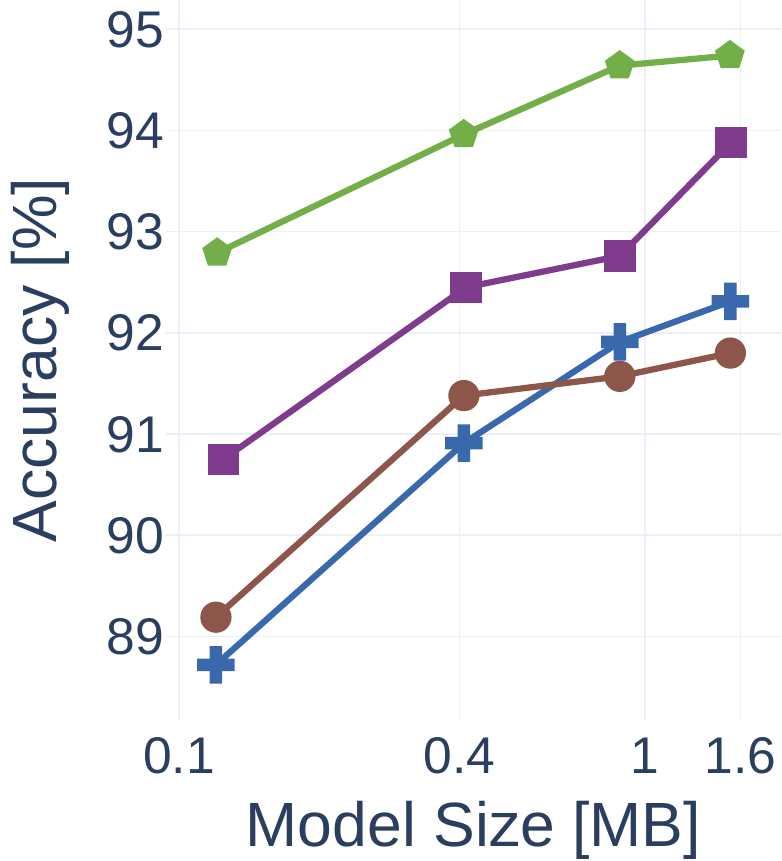}
\caption{Comparison of \reds with Slimmable Subnets~\citep{yu2018slimmable} and width pruning methods (\ie SNIP~\citep{lee2018snip} and magnitude pruning) for subnetworks derived from pre-trained \dscnn models of sizes S and L, evaluated on the Google Speech Commands dataset. The two plots on the left depict test accuracy versus inference time for each subnetwork on the Arduino Nano 33 BLE IoT platform, while the two plots on the right show test accuracy versus model size. \reds consistently outperforms Slimmable Subnets, SNIP, and magnitude pruning, achieving superior accuracy–inference time and accuracy–size trade-offs.}
\label{fig:reds_vs_width_prunings_slimmable}
\end{figure*}

\section{\reds Performance}
\label{sec:reds:performance}
\subsection{Evaluation setup}
We evaluate the performance of \reds, which comprises four nested subnetworks obtained by multiplying the full model MACs (100\%) by the predefined constraints percentages (25\%, 50\%, and 75\%), on three datasets: Google Speech Commands~\citep{warden2019SC} (\googlespeech), Fashion-MNIST~\citep{xiao2017fashion} and CIFAR-10~\citep{cifar100}. In addition, we evaluate four subnetworks on \googlespeech whose MACs budgets correspond to width-pruning ratios of 0.25, 0.5 and 0.75. Moreover, on the Visual Wake Words (\visualwake) dataset~\citep{chowdhery2019visual}, we evaluate two nested subnetworks  under constraints of 50\% of MACs and 250\,KB peak memory usage. We also evaluate three nested subnetworks on the ImageNet-1K dataset~\citep{deng2009imagenet}. 
We measure peak memory usage using the open-source \textit{tflite-tools}~\citep{liberis2021munas}, assuming that the temporary buffer space serves both input and output buffers~\citep{chowdhery2019visual}. MACs counts are obtained via the \textit{tfprof} Tensorflow  profiler. 

\reds is evaluated on: three pre-trained \googlespeech architectures (\dnn, \cnn, and \dscnn) of two sizes (S and L) each introduced in~\citet{zhang2017hello}, an \imagenet~\citep{deng2009imagenet} pretrained \mobilenet~\citep{howard2017mobilenets} architecture fine-tuned on the \visualwake dataset; and a \vit-Base~\citep{dosovitskiy2020image} model trained on \imagenet. For \cifar and \fmnist, we test \reds on two pre-trained, size S \dscnn architectures. The \dnn comprises a two layers fully-connected architecture with layer widths of 144 (S) and 436 (L). The \cnn comprises an architecture with two convolutional layers followed by three fully-connected layers. For the size S model, the convolutional layers have widths of 28 and 30 for the S architecture; for size L, widths of 60 and 76. Both \dscnn and \mobilenet comprise a standard convolutional layer followed by several depth-wise and point-wise convolutional blocks (see \figref{fig:REDS_Convolution_splits.png}). \dscnn has four blocks in the size S network and five in the size L, with layer widths of 64 and 276 respectively. MobileNetV1 includes 13 blocks, beginning with an initial layer width of 32 that progressively increases to 1024 with the model's depth. \vit-Base consists of a patch‐embedding layer followed by 12 transformer blocks, each comprising a multi-head self-attention sub-layer with 12 heads and a multilayer perceptron sub-layer with a hidden dimension of 3072 neurons. Each convolutional layer in the \cnn, \dscnn, and \mobilenet architectures is followed by batch normalization~\citep{ioffe2015batch}. In \vit-Base, each multi-head self-attention and multi-layer perceptron sub-layer is preceded by layer normalization~\citep{ba2016layer}.

\tabref{tab:training-hyperparameters} summarizes the hyper-parameters used to fine-tune the different \dscnn networks taken from~\citet{zhang2017hello}. For \vit we followed the hyperparameters used in~\citet{haberer2024hydravit}. We used the TensorFlow~\citep{abadi2016tensorflow} version 2.12. All models were trained on a workstation with 16 NVIDIA Tesla K80 GPUs and a workstation with 60 NVIDIA A100 GPUs and 564 AMD Ryzen 3 CPUs. To conduct all our experiments and compute the baselines, we trained and optimized over 100 models.

\begin{figure}[t]
    \centering
    \includegraphics[width=0.25\linewidth]{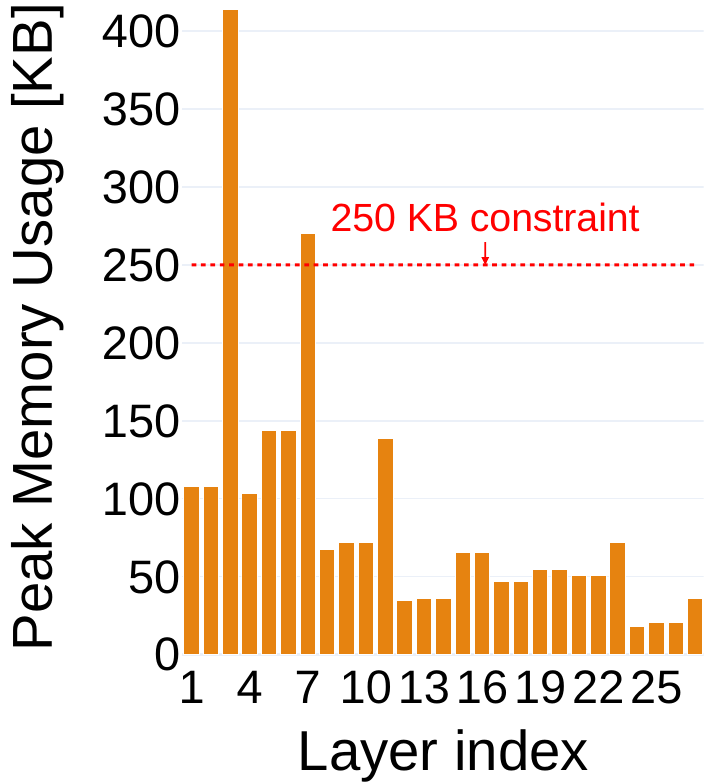}
    \includegraphics[width=0.25\linewidth]{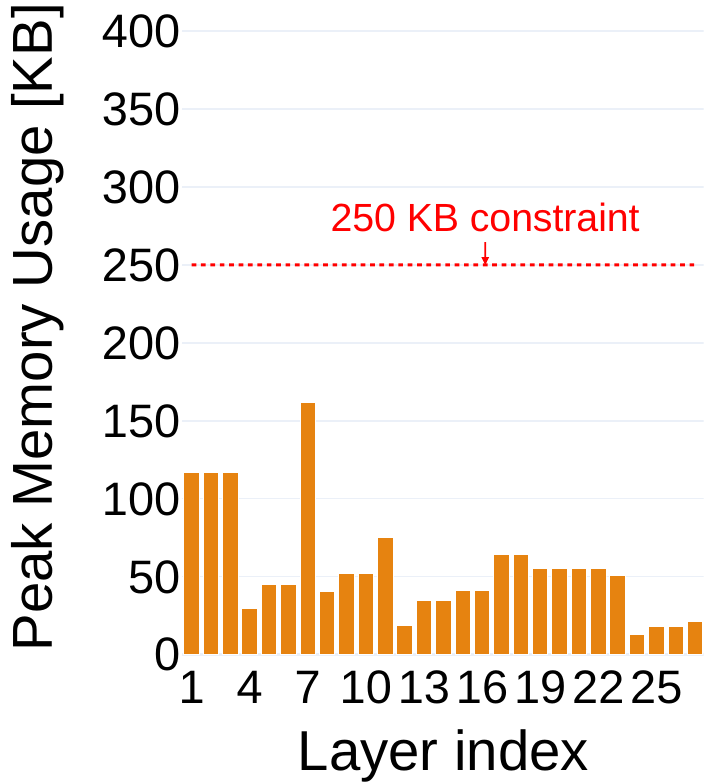}
    \caption{Peak memory usage analysis of \mobilenet processed by the \reds knapsack \bottomup heuristic. Placing constraints on MACs (left), and on both MACs and peak memory usage (right). 
    }
    \label{fig:knapsack_only_macs_vs_macs_peak_mem_constraints}
\end{figure}

\begin{table*}[t]
\caption{Training hyper-parameters.}
\label{tab:training-hyperparameters}
\centering
    	\resizebox{0.48\textwidth}{!}{\begin{tabular}{ccccc}
    		\toprule
    		\textbf{\textbf{Hyper-parameter}} & \textbf{DNN (S/L)}& \textbf{CNN (S/L)}& \textbf{\dscnn (S/L)} \\ 
            \midrule
            Batch size       & \multicolumn{3}{c}{100} \\ 
            Training epochs  & 75  & 75 & 250 \\
            Loss function    & \multicolumn{3}{c}{Cross-entropy} \\ 
            Optimizer        & \multicolumn{3}{c}{Adam} & \\
            LR scheduler & \multicolumn{3}{c}{PiecewiseConstantDecay} \\
            LR    & $10^{-3}$, $10^{-4}$ & $5\cdot10^{-4}$, $10^{-4}$, $2\cdot10^{-5}$ & $10^{-3}$, $10^{-4}$ \\
            LR steps     & $1.5\cdot10^{4}$, & $2\cdot10^{4}$, $10^{4}$, $10^{4}$ & $10^{4}$, $10^{4}$, $10^{4}$ \\
                        & $3\cdot10^{3}$ & \\
            \bottomrule
    	\end{tabular}
        }
\end{table*}

\subsection{\reds empirical evaluation}
\begin{figure*}[!t]
\centering
    \includegraphics[width=0.25\linewidth]{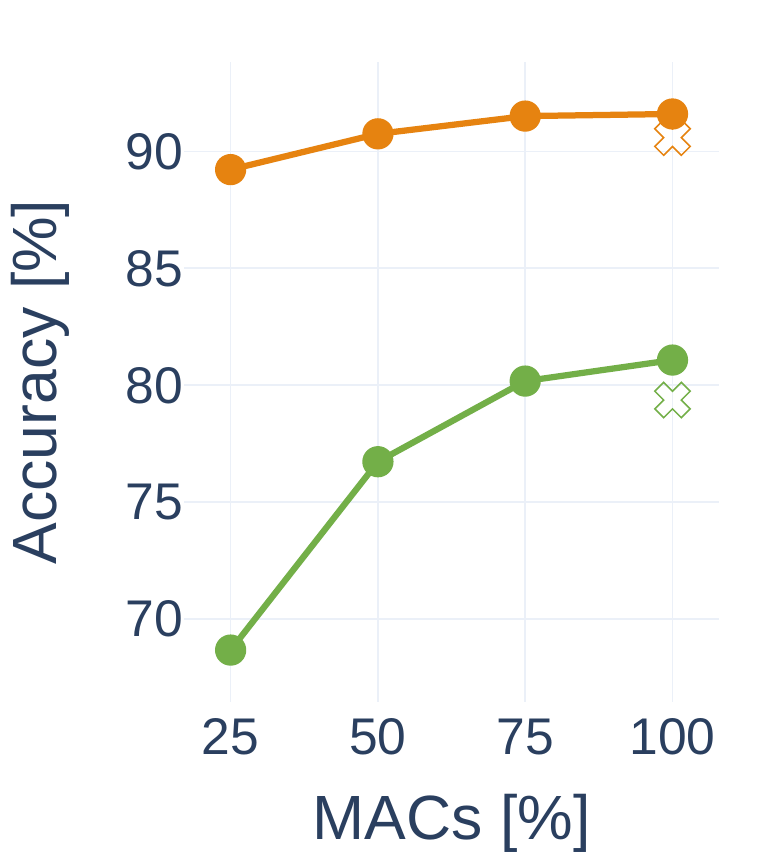} 
    \includegraphics[width=0.25\linewidth]{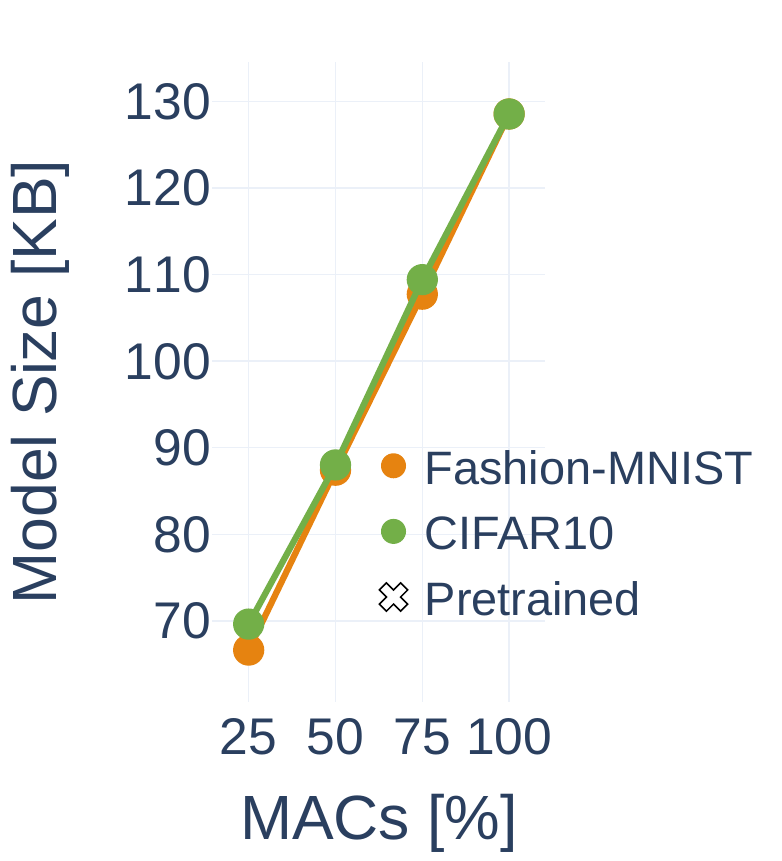}     
    \caption{Analysis of the \bottomup knapsack subnetworks from \dscnn S network, pre-trained on \fmnist~\cite{xiao2017fashion} and \cifar~\cite{cifar100}. 
    \reds supports vision data domain accuracy without degradation of the pre-trained model.} 
    \label{fig:fashionmnist_cifar10_analysis}
\end{figure*}

\begin{figure}[t]
\centering
    \includegraphics[width=0.25\linewidth]{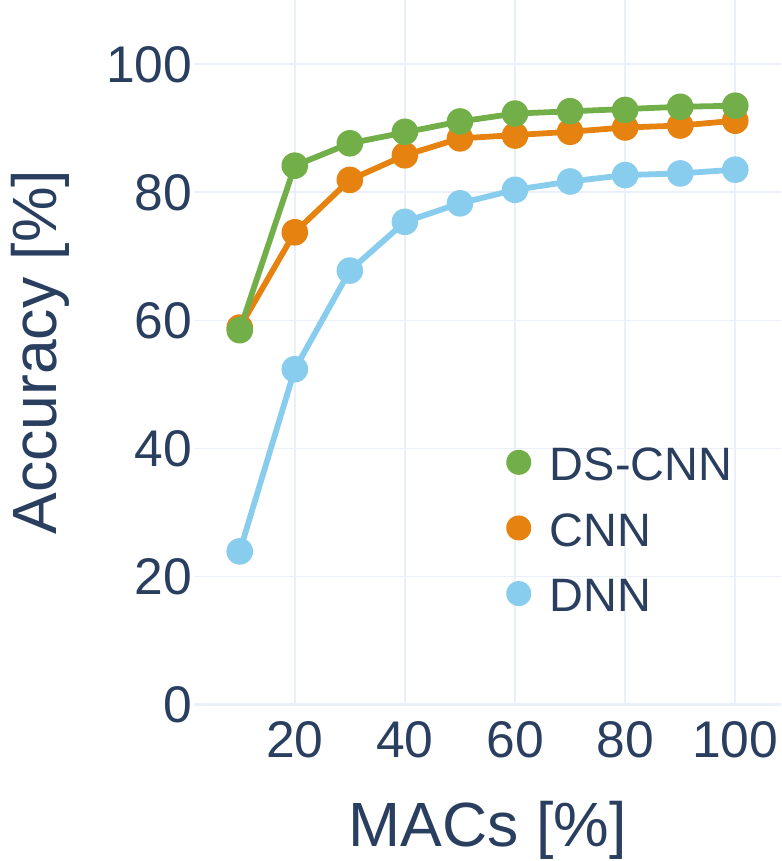} 
    \includegraphics[width=0.25\linewidth]{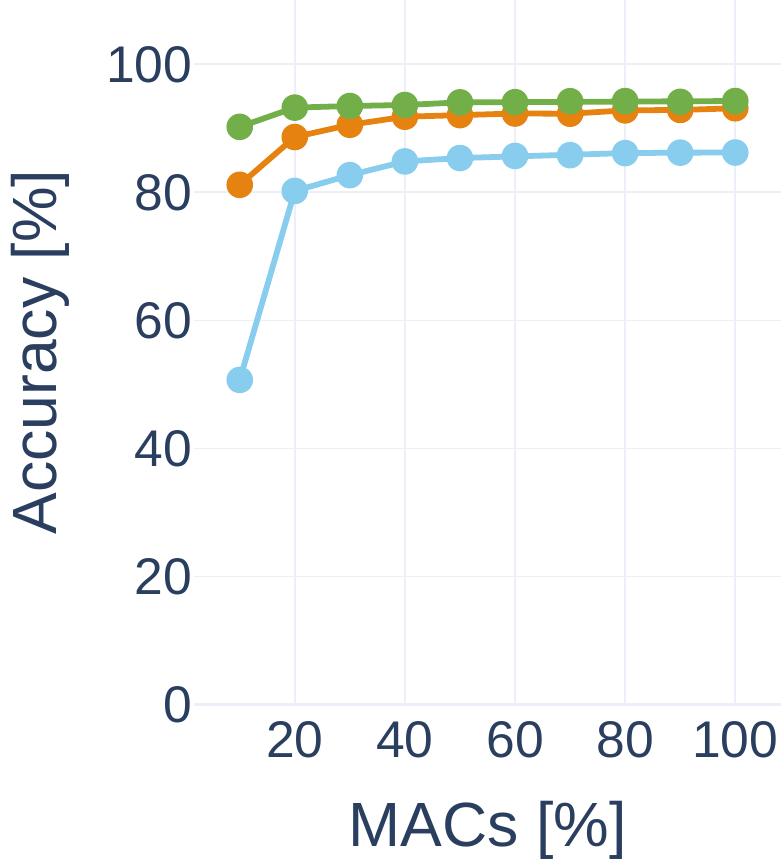}     
    \caption{\reds size S (left) and L (right) \bottomup architectures with ten subnetworks. A larger number of subnetworks does not degrade the accuracy of \reds.} 
    \label{fig:granularity_10subnetworks_configuration}
\end{figure}

The main body of the paper presents the results for \dscnn architectures (S and L) on \googlespeech, \cifar, \fmnist, \mobilenet (0.25x) on \visualwake and \vit on \imagenet. The additional results for other architectures are available in the supplementary extended appendix.
The \reds structure for the considered datasets used in the main paper comprises only four nested subnetworks; however a larger number of subnetworks does not degrade the accuracy of \reds fine-tuned on \googlespeech  (see \figref{fig:granularity_10subnetworks_configuration}). The results in \figref{fig:fashionmnist_cifar10_analysis} show \reds performance using \dscnn architecture of size S on \fmnist and \cifar. \bottomup heuristic was used to obtain the results. \reds supports a different data domain without degrading the accuracy of the pre-trained model, reported in the header row. Curiously, jointly training the subnetworks improves the performance of the pre-trained 100\% MACs subnetwork, similarly as reported in~\citet{yu2018slimmable}.

The effectiveness of \reds for MCUs and mobile architectures is shown in \tabref{tab:finetuningvsfulltraining_ds_cnn} and \tabref{tab:performance_comparison_visual_wake} respectively.
Training from scratch shows the accuracy achieved by S and L \dscnn architectures when trained in isolation, \ie not as part of the \reds nested structure. The next two columns show the resulting \reds performance when the solution is found by the \bottomup and \topdown heuristics, respectively. Finally, the forth column reports \reds test set accuracy achieved by each \reds subnetwork trained from scratch, yet following the structure obtained by the \bottomup heuristics. 

\begin{table*}[t]
\begin{center}
    \caption{Test set accuracy [\%] of training S and L depth-wise separable convolutional architectures (\dscnn) from \citet{zhang2017hello}: Training a network of each size from scratch ("Scratch"), conversion from a pre-trained network using "Knapsack \bottomup" and "Knapsack TD", and training \reds structure from scratch ("\reds training"). Reported results from three independent runs. The accuracy of each 100\,\% network reported in \citet{zhang2017hello} is listed in the header row.}
    \label{tab:finetuningvsfulltraining_ds_cnn}
\end{center}
\resizebox{\textwidth}{!}{
\begin{tabular}{c|cccc|cccc}
\toprule
\textbf{MACs}
& \multicolumn{4}{c|}{\textbf{Small (S) - Accuracy 94.96}}  
& \multicolumn{4}{c}{\textbf{Large (L) - Accuracy 95.1}} \\ 
& Scratch & Knapsack \bottomup & Knapsack \topdown & \reds training
& Scratch & Knapsack \bottomup & Knapsack \topdown & \reds training\\
\midrule
100\%                  & 93.83$\pm 0.22$  & 93.38$\pm0.45$  & 93.34$\pm0.21$ & 93.19 $\pm0.18$ & 94.87 $\pm0.33$& 94.46$\pm0.08$ & 94.35 $\pm0.21$ & 94.05 $\pm0.14$ \\
75\%                   & 93.37$\pm 0.34$  & 93.18$\pm0.21$   & 93.03$\pm0.26$ & 92.85 $\pm0.47$ & 94.27 $\pm0.08$& 94.32$\pm0.13$ & 94.18 $\pm0.17$ & 93.76 $\pm0.02$ \\
50\%                   & 93.41$\pm 0.67$  & 92.12$\pm0.24$   & 92.26$\pm0.50$  & 91.50  $\pm0.10 $ & 94.11 $\pm0.26$& 94.17$\pm0.01$ & 94.00    $\pm0.25$ & 93.08 $\pm0.06$ \\
25\%                   & 91.46$\pm 0.80$   & 89.64$\pm0.76$   & 88.59$\pm1.69$ & 85.14 $\pm1.07$ & 93.80  $\pm0.14$& 93.20$\pm0.19$  & 93.17 $\pm0.73$ & 92.11 $\pm0.42$ \\
\bottomrule
\end{tabular}}
\end{table*}

\begin{figure}[!t]
\centering
    \includegraphics[width=0.25\linewidth]{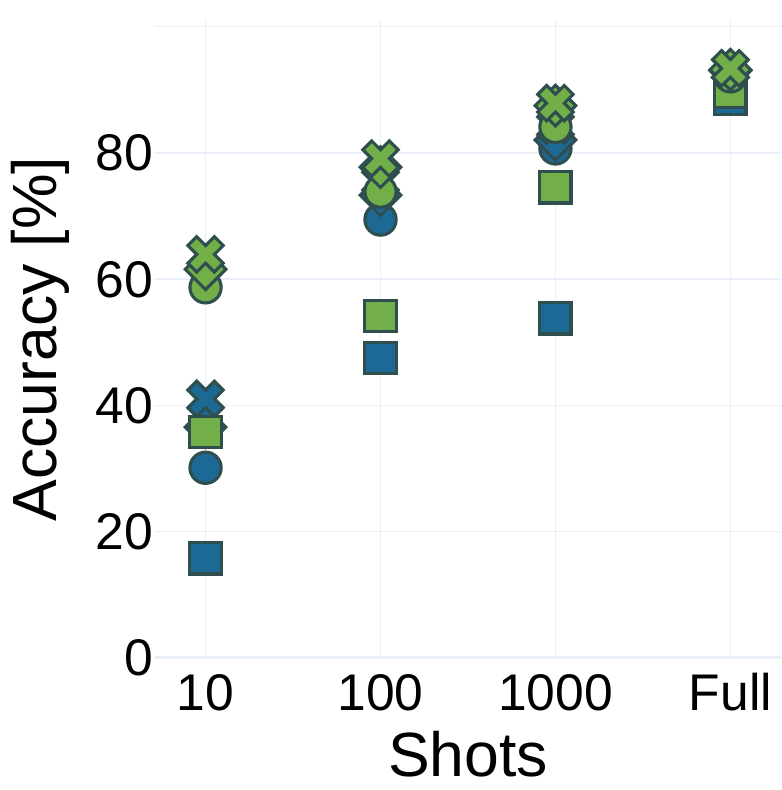} 
    \includegraphics[width=0.25\linewidth]{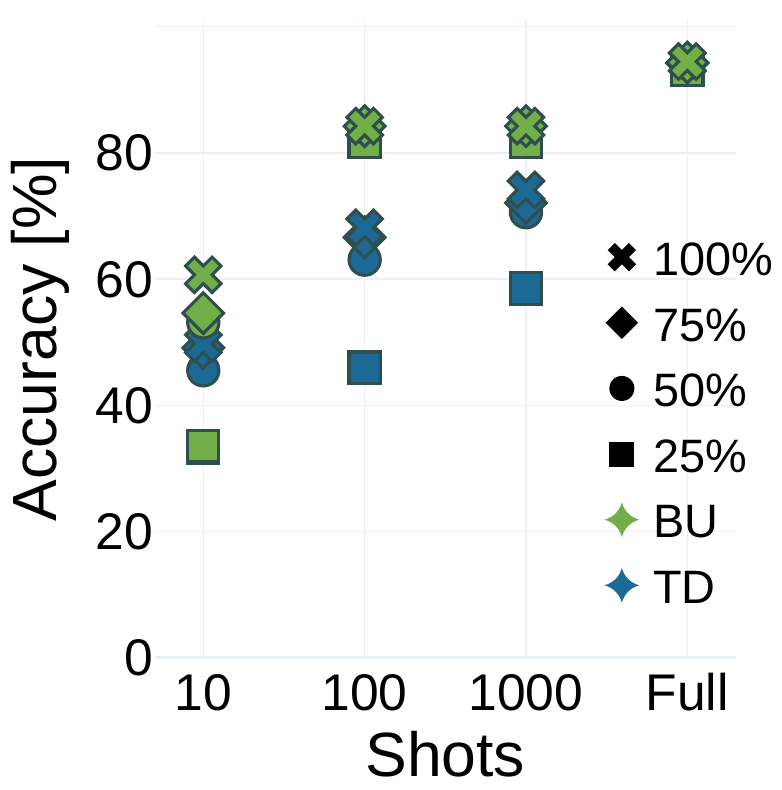}     
    \caption{\dscnn S and L few-shots finetuning on \googlespeech~\cite{warden2019SC} with MACs percentages as constraints. The subnetworks obtained from the \bottomup Knapsack formulation exhibit faster recovery in accuracy compared to the \topdown Knapsack ones.} 
    \label{fig:bu_vs_tp_few_shots_analysis}
\end{figure}

We observe that the accuracies of each submodel for the same percentage of MACs are similar, even for small S architectures with only 25\,\% of MACs. The accuracies of the full models, highlighted in bold in the top row, are computed using pre-trained models from \citet{zhang2017hello} and closely match our experimental results. The performance difference between the \bottomup and \topdown heuristics is largely insignificant due to the fine-tuning step with extensive training data. Even though the models are overparameterized and in all cases achieve high performance, low-capacity subnetworks S sliced by the bottom-up approach yield minor performance gains. However in the low-data regime, where only few samples per class are available for fine-tuning, the performance differences between \bottomup and \topdown heuristics are remarkable. \figref{fig:bu_vs_tp_few_shots_analysis} shows the performance when few-shot learning is applied to fine-tune \reds. Our empirical findings confirm a consistently superior performance of the \bottomup heuristic compared to the \topdown alternative. The differences diminish as more samples are used for fine-tuning~\citep{entezari2023role} and there is barely any difference if full-finetuning is applied using all available training data.

\begin{table}[t]
    \begin{center}
    \caption{Comparative analysis of \reds and \citet{chowdhery2019visual} on MobileNet v1 on \visualwake pruned down to 0.25\%. \reds yields 27\% lower peak memory usage for 0.9\% lower accuracy.}
    \label{tab:performance_comparison_visual_wake}
    \resizebox{0.48\textwidth}{!}{
    \begin{tabular}{l|cc}
    \toprule
     & Peak-Memory-Usage & Test Accuracy \\
    \midrule
    \reds                   & 162 KB & 74.8\% \\
    MobileNet v1 (0.25x) & 223 KB & 75.71\% \\
    \bottomrule
    \end{tabular}}
    \end{center}
\end{table}

For the \mobilenet architecture, we compare \reds to MobileNet v1 (0.25x) using peak memory constraint of 250 KB and the test accuracy previously defined in~\citet{chowdhery2019visual}. \reds knapsack \bottomup heuristic achieves 27\% lower peak memory usage for only 0.9\% lower accuracy (see  \tabref{tab:performance_comparison_visual_wake}). This validates the efficacy of the knapsack \bottomup approach in constraining peak memory usage within the subnetworks' architecture. Furthermore, imposing a peak memory usage constraint is essential to facilitate the deployment of the \reds within the confines of the available device RAM (see \figref{fig:knapsack_only_macs_vs_macs_peak_mem_constraints}).

\figref{fig:reds_vs_width_prunings_slimmable} reports the performance of \reds for accuracy-inference time (left) and accuracy-model size (right) for \dscnn S and L on the \googlespeech dataset, measured on an Arduino Nano 33 BLE. \reds subnetworks form a strictly superior Pareto frontier: at any given latency or model size, they achieve higher test accuracy than magnitude pruning, SNIP and Slimmable Subnets. These gains stem from \reds' structured sparsity design, which selects contiguous blocks of filters and encodes each subnetwork via simple slicing points. Thereby eliminating per‐inference binary‐mask lookups and from its single set of batch‐normalization parameters permits compile‐time fusion of batch normalization parameters into the convolution weights. In contrast, width pruning methods incur runtime mask‐application overhead; and Slimmable Subnets store duplicate batch normalization parameters for each width (\ie precluding batch normalization fusion), resulting in storage and inference time overheads. 

\tabref{tab:vit_subnet_acc} presents the Top-1 accuracy of nested \vit-Base subnetworks trained from scratch obtained by considering two MACs constraints (4.60 GMACs and 1.25 GMACs) on ImageNet-1K validation set. Although HydraViT~\citep{haberer2024hydravit} slightly outperforms \reds at the full model configuration (80.45\,\% vs 79.88\,\%), \reds achieves superior accuracy at both the medium (78.42\,\% vs 78.40\,\% and 77.79\,\%) and low (68.21\,\% vs 67.34\,\%  and 66.64\,\%) subnetworks. This improvement is due to \reds's iterative \vit knapsack formulation, which allocates attention heads and projection dimensions to maximize the importance score within fixed MACs constraints. In contrast, HydraViT’s fixed head-partitioning and SortedNet’s uniform scaling do not adaptively prioritize the most informative components as MACs constraints become limited. Consequently, \reds achieves consistently higher accuracy–compute trade-offs without altering the underlying contiguous storage of subnetwork parameters.

\begin{figure}[t]
    \centering
    \includegraphics[width=0.25\linewidth]{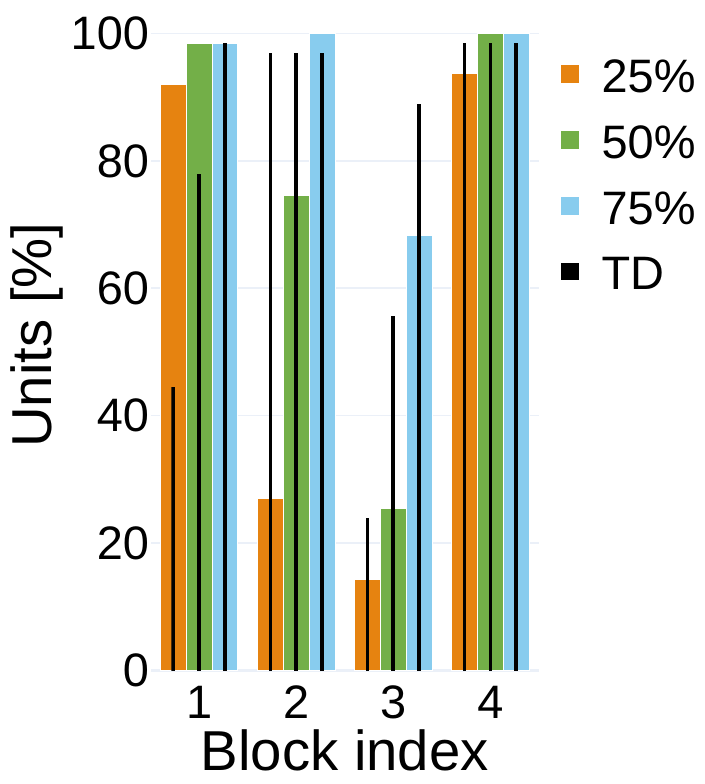}
    \includegraphics[width=0.25\linewidth]{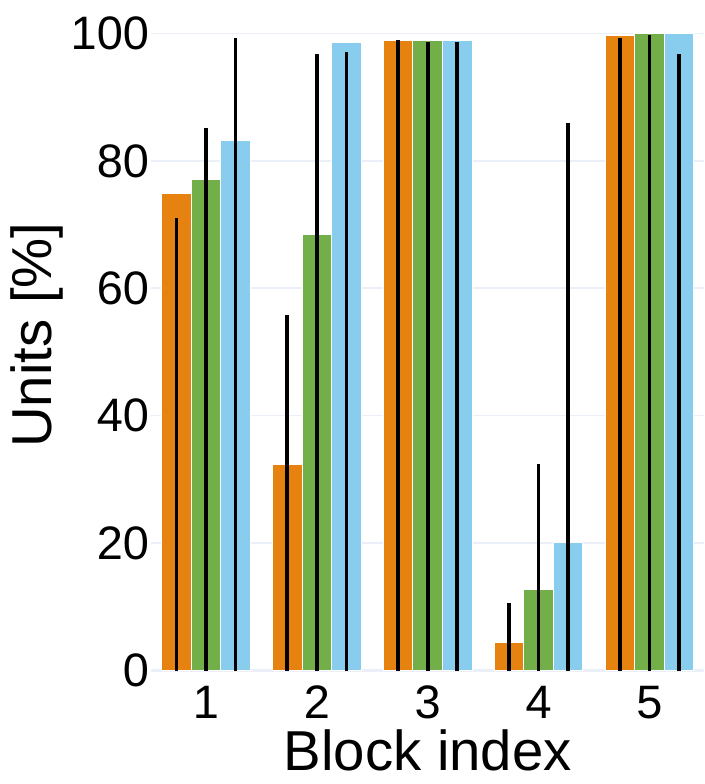}
    \caption{Analysis of the subnetworks' structured sparsity obtained by the knapsack \bottomup and \topdown heuristics (\secref{sec:iterative_knapsack}) with MACs percentages as constraints. Left to right: \dscnn S and \dscnn L. The slicing point of each subnetwork is visualized with a different color. The results of the \topdown heuristic are reported in black.}
    \label{fig:structural_pruning_models:dscnn}
\end{figure}

\figref{fig:structural_pruning_models:dscnn} visualizes differences between \reds structures obtained with the \bottomup and \topdown heuristics when using the same initially pre-trained \dscnn models. Colored bars show the number of filters kept by each nested model in every layer by \bottomup. Overlaid black bars correspond to the number of kept \topdown filters for the same layer and the number of MACs. The plot reveals consistent differences in each layer as we move from parent to child submodels. The distribution of computational units across subnetworks in each layer depends on the model architecture (see supplementary extended appendix \ref{appendix:further:eval}).

\begin{table}[t!]
    \begin{center}
        \caption{Comparison of Top-1 accuracy for nested \vit-Base subnetworks trained from scratch under two MAC constraints. The \reds subnetworks, derived from the \vit-Base \topdown knapsack formulation, match the MAC counts of the standard N-head \vit-Base models (4.60 and 1.25 GMACs) while employing larger embedding dimensions. For the full model (\ie~17.56 GMACs), HydraViT achieves 80.45\% accuracy, surpassing the 79.88\% of \reds. However, at the medium (4.40 GMACs) and low (1.24 GMACs) MACs constraints, \reds outperforms both HydraViT and SortedNet.}
        \label{tab:vit_subnet_acc}
        \begin{tabular}{c|c|c|c}
        \toprule
        \textbf{Method} & \textbf{Dim} & \textbf{MACs [G]} & \textbf{Acc.\ [\%]} \\
                        &    &                   & (scratch)         \\
        \midrule
        HydraViT~\citep{haberer2024hydravit}        & 768          & 17.56             & \textbf{80.45}             \\
                        & 384          & 4.60              & 78.40             \\
                        & 192          & 1.25              & 67.34             \\
        \midrule
        REDS            & 768          & 17.56             & 79.88             \\
                        & 512          & 4.40              & \textbf{78.42}             \\
                        & 256          & 1.24              &  \textbf{68.21}             \\
        \midrule
        SortedNet~\citep{valipour2023sortednet}       & 768          & 17.56             & 79.71             \\
                        & 384          & 4.60              & 77.79             \\
                        & 192          & 1.25              & 66.64             \\
        \bottomrule
        \end{tabular}
    \end{center}
\end{table}

\section{REDS Optimization for Caches}
\label{sec:cache}

Embedded ML frameworks like TFLite typically store model weight matrices in row-major order.
This means that each row of a weight matrix is stored contiguously in memory. 
Without the use of subnetworks these weights are also accessed in a contiguous fashion.
However, when using subnetworks for model inference, some neurons and their weights are omitted, resulting in non-contiguous memory access.
This effect is illustrated in Figure~\ref{fig:REDS_Dense_splits}.
We now show how by adapting the computational graph at compile time, we are able to optimize the computation of \reds subnetworks for devices with a cache memory architecture.

\subsection{Row-major and column-major stores}
The calculation of a fully-connected neural network layer during the forward pass is a matrix multiplication $H = \sigma(X_{[m\times b]}^T\cdot W_{[m\times n]})$, where $x$ is an input matrix of shape $m\times b$ (input samples $\times$ batch size), $W$ the layer weights matrix of shape $m\times n$ (input samples $\times$ number of neurons) and $\sigma$ the activation function.
For simplicity we omit the typically used bias term.
During matrix multiplication, like it is for instance implemented in TFLite, an inner loop multiplies and sums each element of row $x_i$ of $X^T$ with column $w_j$ of $W$.
This column-wise access of row-major ordered weights $w\in W$ may lead to consecutive reads from memory addresses, which are not located close-by depending on the size of the matrix and the use of the \reds subnetworks.
However, this can easily be circumvented by changing the matrix multiplication involving a transposed $W$, \ie $\sigma(X^T \cdot W) = \sigma((W^T \cdot X)^T)$, which ensures contiguous memory accesses of the weights.
Note that the additional transposition of the resulting matrix introduces no additional computations if the input matrix is a single sample, \ie $X$ has size $[m\times 1]$.
Both $X_{[m\times 1]}$ and $X_{[m\times 1]}^T$ are stored equally in memory and, thus, the transposition is redundant in this case.
Let us consider the simple example of $X_{[2\times 2]}$ and $W_{[2\times 3]}$ both stored in row-major mode.
In the basic matrix multiplication cases $X^T\cdot W$, the order of the relative memory access locations of weights $w$ is $0 \rightarrow 3 \rightarrow 1 \rightarrow 4 \rightarrow 2 \rightarrow 5 \rightarrow 0 \rightarrow 3 \rightarrow \cdot\cdot\cdot$.
In the optimized cases $(W^T\cdot X)^T$ where $W^T$ is also stored in column-major form, the order of weights accesses is $0 \rightarrow 1 \rightarrow 0 \rightarrow 1 \rightarrow 0 \rightarrow 1 \rightarrow 2 \rightarrow 3 \rightarrow \cdot\cdot\cdot$.
Depending on the memory and cache setup of a device, this highlights the potential for exploiting cache systems by simply changing the matrix multiplications to the optimized transposed form.

\begin{table}[t!]
    \begin{center}
        \caption{Speed-up and cache-hit rate for different matrix data types. There is no notable difference across the different splits 25, 50, 75 and 100\%.}
        \label{tab:cache_stats}
        \resizebox{0.48\textwidth}{!}{
        \begin{tabular}{l|cccc}
        \toprule
        & \texttt{float} & \texttt{uint8} & \texttt{uint16} & \texttt{uint32} \\
        \midrule
        Speed-up & 12\% & 58.6\% & 54.5\% & 47.4\% \\
        \midrule
        Cache-Hit-Rate Baseline & 99.1\% & 90.9\% & 91.7\% & 90.9\%  \\
        Cache-Hit-Rate Optimized & 99.7\% & 99.4\% & 98.9\% & 97.6\% \\
        \bottomrule
        \end{tabular}}
    \end{center}
\end{table}

\begin{figure}[t!]
    \centering
    \includegraphics[width=0.25\linewidth]{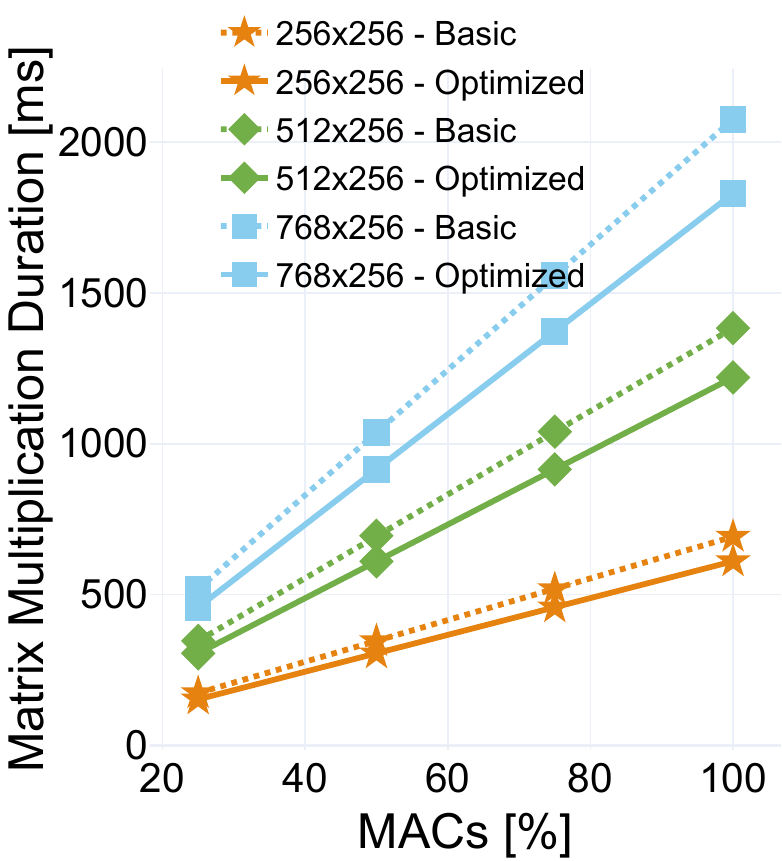}
    \includegraphics[width=0.25\linewidth]{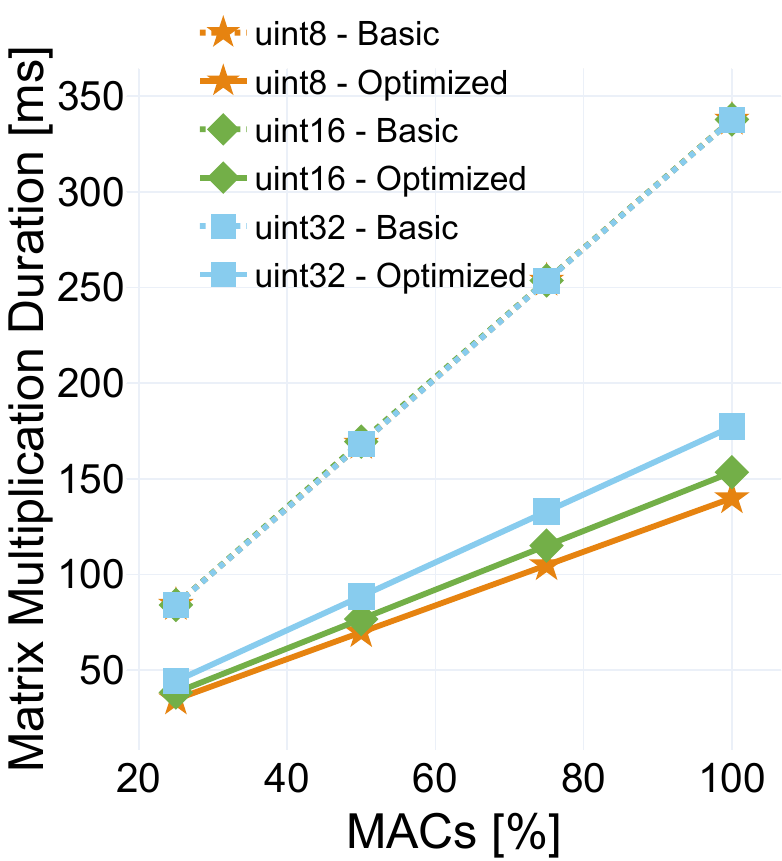}
    \caption{Duration of the matrix multiplication using differently sized weight matrices of floats (left) and different unsigned integer precisions for a 512 x 256 weight matrix (right) at different weight matrix slicing points. The cache optimization shows a clear computational time reduction for all the different test scenarios.}
    \label{fig:cache_opt_matmul_duration}
\end{figure}

\subsection{Optimizing \reds computational graph}
The optimization proposed above can be implemented by assuring that each matrix multiplication within a computational graph of each submodel follows the cache optimized access pattern and the corresponding weight matrices are stored in column-major order in memory.
We now show the effect of using this optimized computational graph by benchmarking different matrix multiplications.
To this end, we use a Raspberry Pi Pico board, which features a RP2040 chip based on a dual-core Arm Cortex-M0+ processor architecture with 264kB RAM and 16MB off-chip flash memory~\cite{rp2040}.
The chip also features execute-in-place (XIP) support, such that the flash memory can be treated like internal memory.
These accesses to flash are cached with a 16kB, two way set-associative cache with 8 byte wide cache lines.

\figref{fig:cache_opt_matmul_duration} shows the cache effect on the duration of matrix multiplications when we use the optimized computation graph compared to the basic one, \ie the default way where W is stored in row-major mode and we calculate $X^T\cdot W$.
We benchmark the matrix multiplication at different splits and use 25\%, 50\%, 75\% and 100\% of the neurons in $W$.
In \figref{fig:cache_opt_matmul_duration} (left) we show the duration of a matrix multiplication with differently shaped weights $W$ times an input matrix $X$ of shape $256\times4$, where the weights and inputs are 32\,Bit wide floating points.
Similarly, \figref{fig:cache_opt_matmul_duration} (right) shows the duration of a weights matrix multiplication of size $512\times256$ with the same $X$, but now the weights and inputs are unsigned integers with different widths, \ie 8, 16 and 32\,Bit. 
We observe in all cases a clear speed-up in matrix multiplication when comparing the optimized and the basic computation approach, which is also evident from Table~\ref{tab:cache_stats}.
For all test cases the cache-hit rate is above 97\% when using optimized matrix multiplications.
Additionally, we see a notable difference between floating point and integer based matrices.
This difference is due to the fact that the RP2040 does not feature any floating point arithmetic support and, thus, the cache effect is diminished due to longer duration of basic multiplications.
Finally, there is no notable difference in speed-up and cache-hit rate between the different splits.

We can therefore conclude that a simple change in the computational graph at compile time, \ie using column-major weight matrices, leads to increased matrix multiplication speeds on devices featuring a cached memory. Cache optimization is thus essential to speed-up inference of nested submodels part of \reds.

\begin{figure*}[t!]
    \centering
        \includegraphics[width=0.24\textwidth]{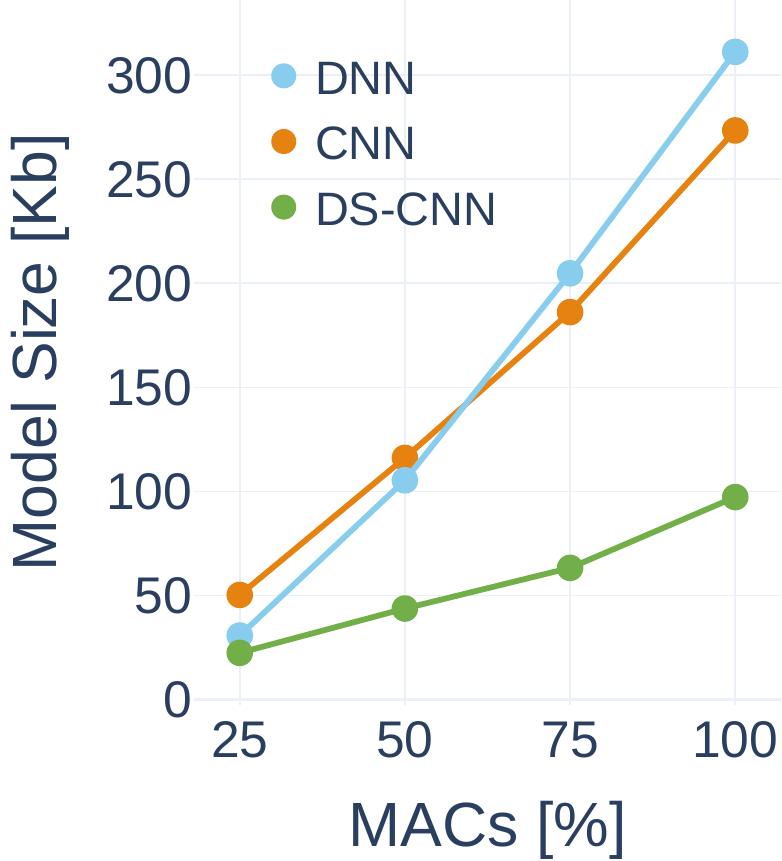}
        \includegraphics[width=0.24\textwidth]{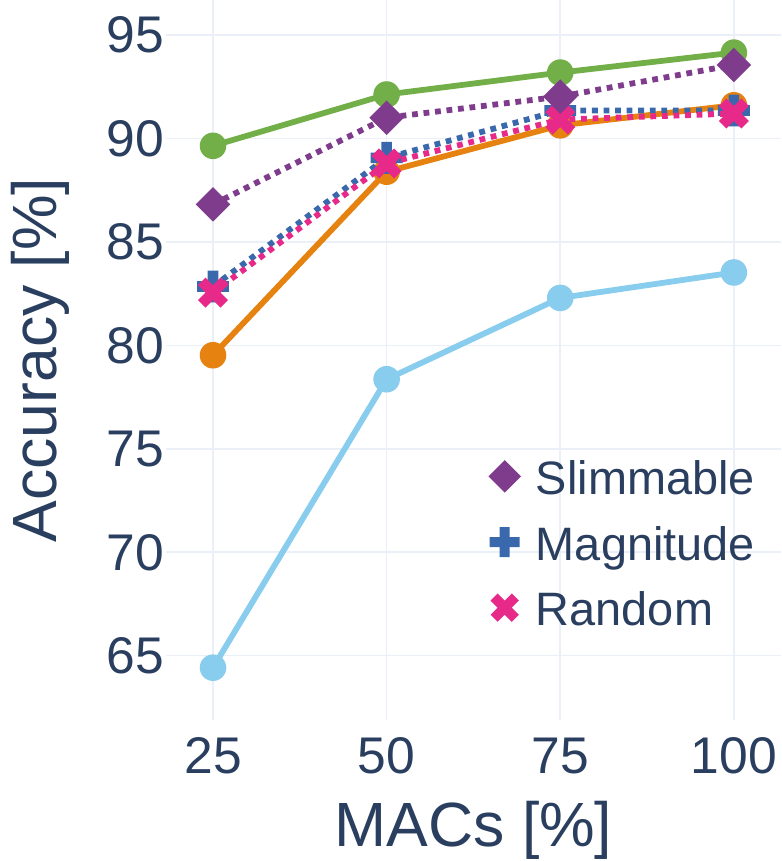}        
        \includegraphics[width=0.24\textwidth]{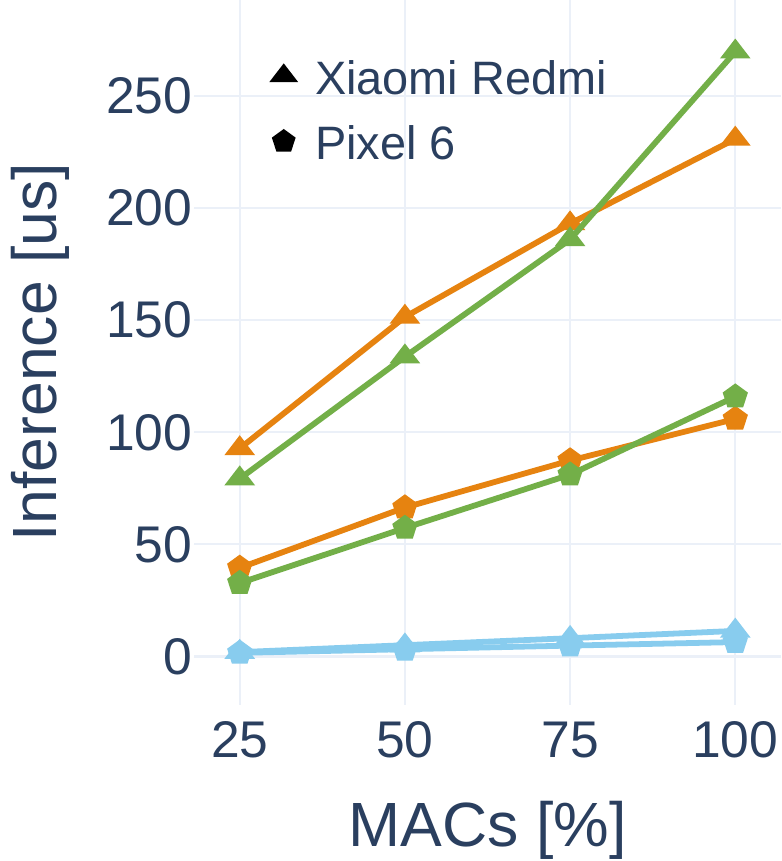}
        \includegraphics[width=0.24\textwidth]{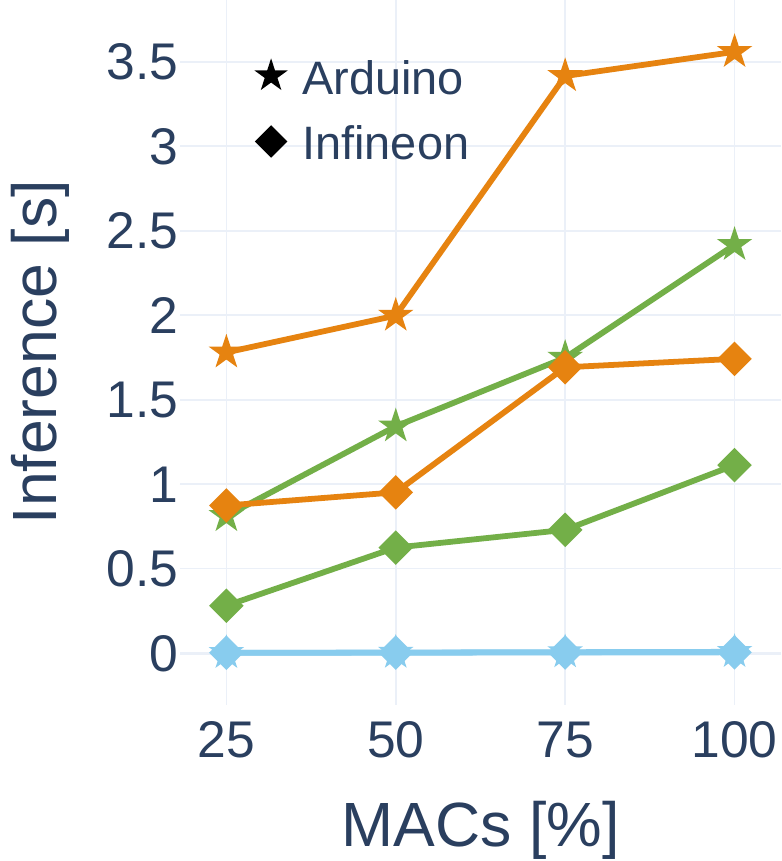}

        \includegraphics[width=0.24\textwidth]{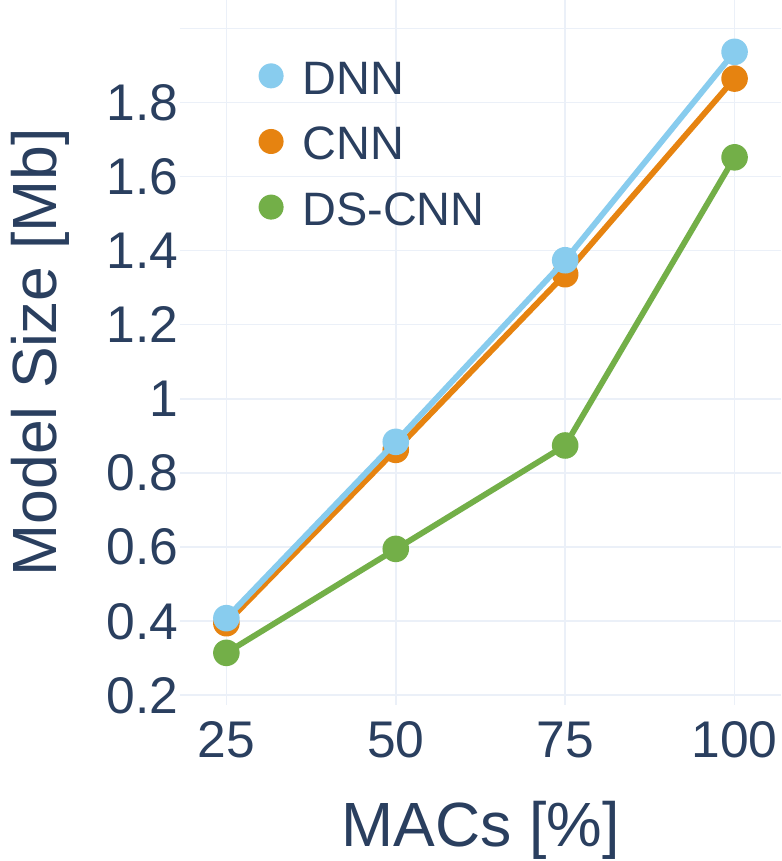}
        \includegraphics[width=0.24\textwidth]{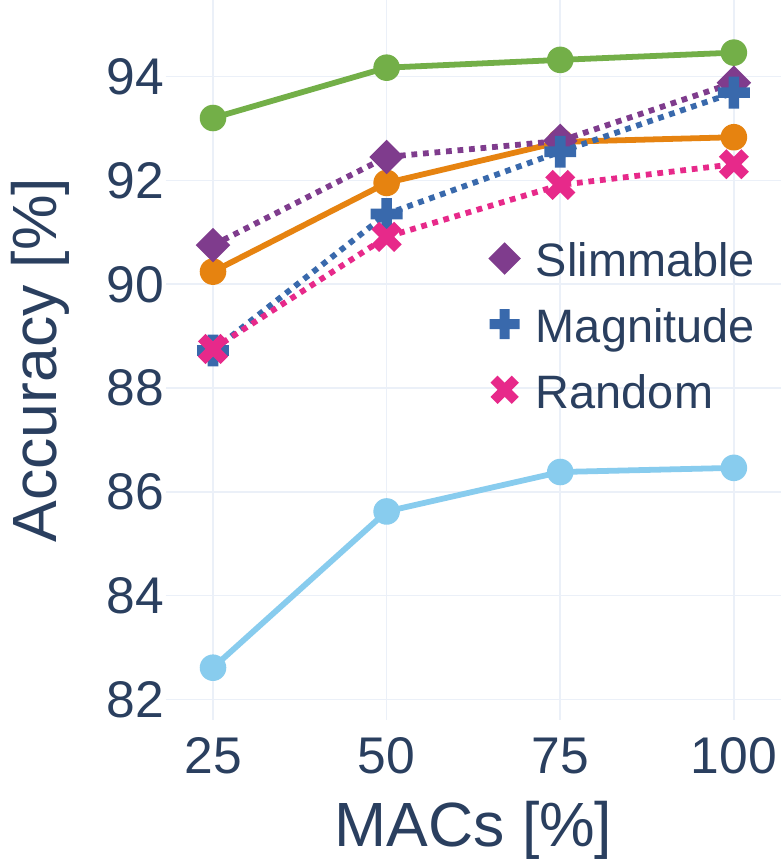}
        \includegraphics[width=0.24\textwidth]{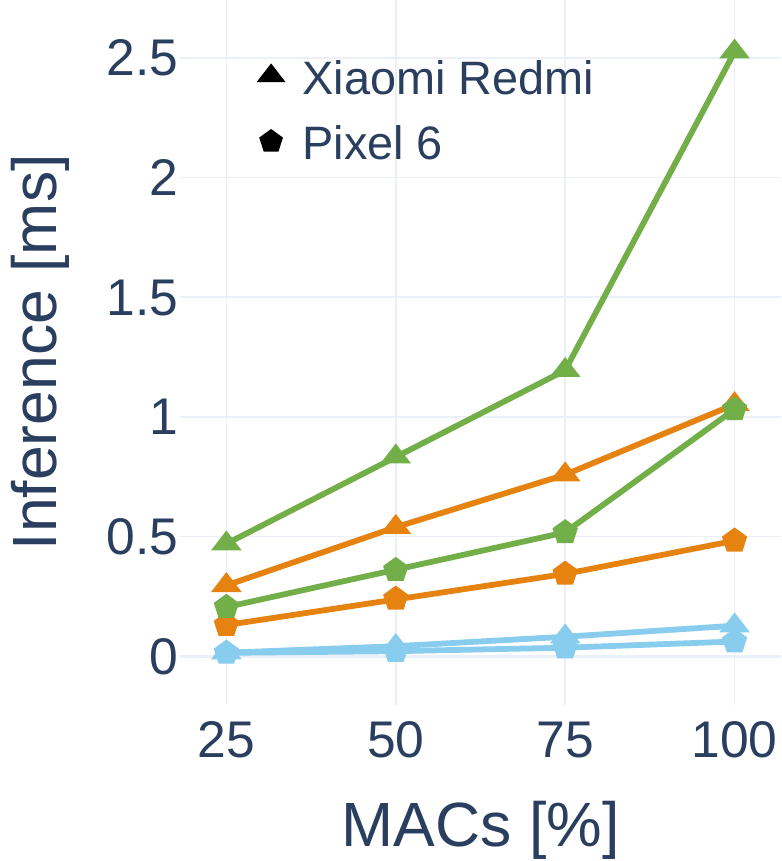}
        \includegraphics[width=0.24\textwidth]{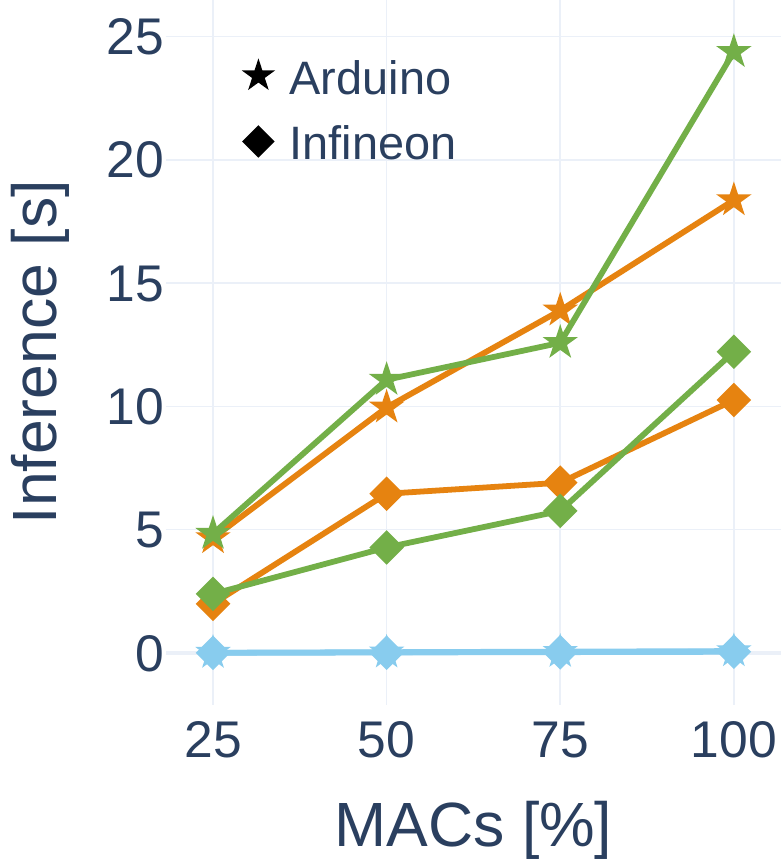}        
    \caption{\reds size S (top row) and L (bottom row) architecture analysis. The two plots on the left show the number of model parameters and model accuracy as a function of MAC percentage in each \reds submodel. The two plots on the right evaluate model inference time on two classes of devices: more powerful platforms comprising Xiaomi Redmi Note 9 Pro (Qualcomm Snapdragon 720G, ARM Cortex-A76), Pixel 6 (Octa-core 2x2.8\,GHz Cortex-X1, 2x2.25\,GHz Cortex-A76, 4x1.8\,GHz Cortex-A55); and low-power IoT platforms including Arduino Nano 33 BLE Sense (nRF52840, ARM Cortex-M4) and Infineon CY8CKIT-062S2.}
    \label{fig:mobile_size}
\end{figure*}

\begin{figure}[t!]
    \centering
    \includegraphics[width=0.25\linewidth]{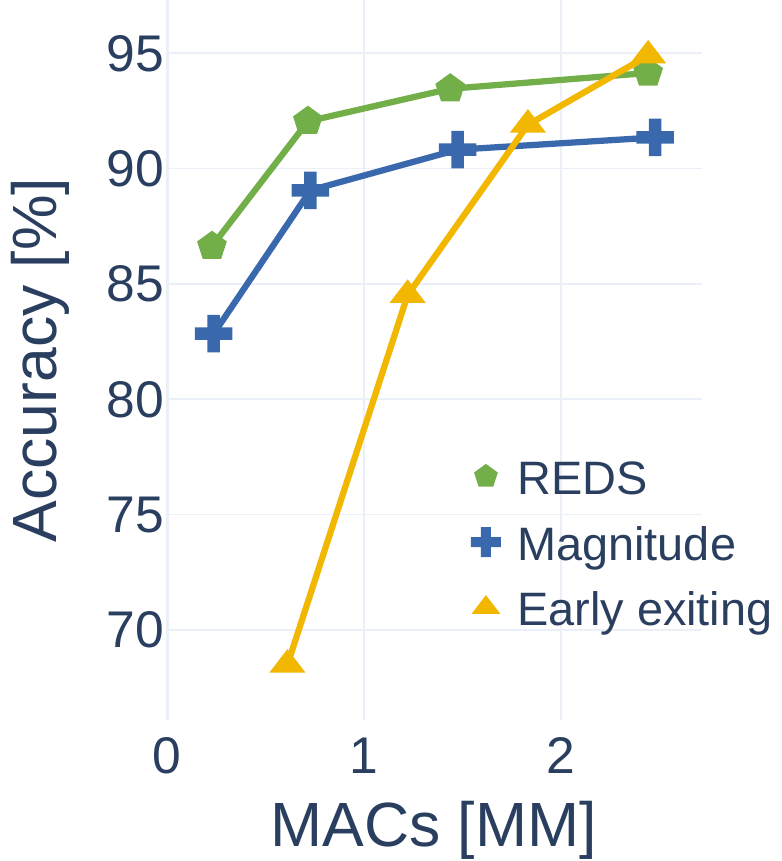} 
    \includegraphics[width=0.25\linewidth]{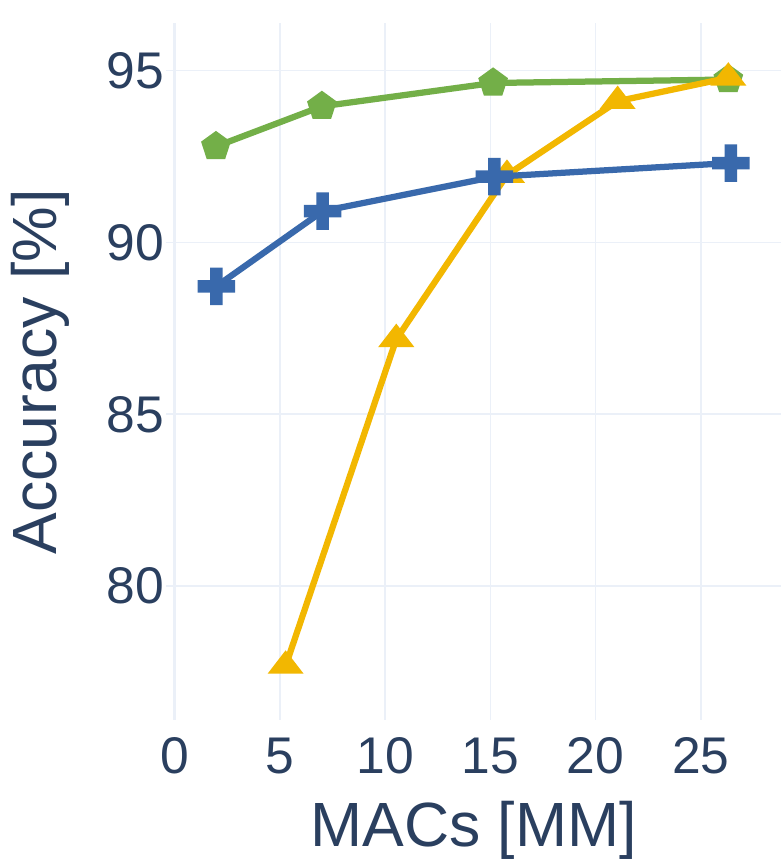}     
    \caption{\dscnn S (left) and L (right) \bottomup comparison of \reds, early exit linear classifiers and magnitude width multiplier pruning subnetworks.} 
    \label{fig:early_exits_ds_cnn_s_l_l1_comparison}
\end{figure}

\section{\reds on Mobile and IoT}
\label{sec:iot}
We evaluate \reds models on two mobile phones, specifically Xiaomi Redmi Note 9 Pro and Google Pixel 6, and on two IoT devices, namely Arduino Nano 33 BLE Sense and Infineon CY8CKIT-062S2. For the mobile phone evaluation, we employed the Google TFLite benchmarking tool, which measures the model's inference time on Android-based mobile devices. On IoT devices we deployed and evaluated the models using the Edge Impulse~\citep{hymel2022edge} platform.
We report the number of parameters, accuracy, and latency on the mobiles and on the IoT devices for the full architectures (100\% MACs) and the subnetworks architecture (75\%, 50\% and 25\% MACs) found by the \bottomup knapsack solvers. The obtained results are reported in \figref{fig:mobile_size} for the \dnn, \cnn and \dscnn architectures S (top row) and L (bottom row) on \googlespeech. All results are averages over three runs. TFLite and TFLMicro lack support for runtime adaptation of model weight tensors. We extended the TFLMicro framework to support \reds out of the box and measure on-device inference and submodel switching times. The code is publicly available\footnote{Link: \url{https://github.com/aschesklave/TFlite-Micro-Cutting}}.

The first left-most column of plots in \figref{fig:mobile_size} shows the relationship between the subnetworks' MACs and the number of model parameters. All curves are close to linear, yet the parameters of these linear relationships are architecture-specific. The \dscnn models have the lowest number of parameters, thanks to their more sophisticated architecture, which is more efficient on our dataset compared to the standard convolutional and dense networks. \dscnn models also yield better accuracy, even for only 25\% of MACs, while \dnn models perform worst, as can be observed in the second column of plots. This result can be attributed to the successful generalization of the iterative knapsack problem to support depth-wise convolutions \secref{appendix:knapsack_variant_two}.

\begin{table}[t!]
    \begin{center}
        \caption{Energy consumption of \reds size S architectures measured by the Power Profiler Kit (PPK2) on Nordic nRF52840. The results are obtained by running an inference pass through each subnetwork with input data of batch size one, while recording the corresponding inference current.} 
        \label{tab:power_profiler_kit_benchmark}
        \begin{tabular}{c|c|c|c}
        \toprule
        \textbf{MACs} & 
        \textbf{\dnn} & 
        \textbf{\cnn} & 
        \textbf{\dscnn} \\
              & [mJ] & [mJ] & [mJ] \\
        \midrule
        100\% & 0.18 & 90.61 & 61 \\
        75\%  & 0.15 & 86.01 & 44.03 \\
        50\%  & 0.07 & 50.3 & 33.81 \\
        25\%  & 0.05 & 44.81 & 20.44 \\
        \bottomrule
        \end{tabular}
    \end{center}
\end{table}

We compare \reds to several works. We consider model optimization by pruning an equal share of computational blocks in each layer~\citep{tan2019efficientnet} (\ie width multiplier pruning). We use the block selection strategy based on the magnitude of the weight (\ie magnitude pruning) used in \citet{cai2020onceforall}, and a random selection. Additionally, we compare to Slimmable Subnets~\citep{yu2018slimmable}, where each layer stores one extra batch norm statistic for each subnetwork, also obtained by width multiplier pruning. We did not compare \reds to DRESS or NestDNN  because neither method’s source code is publicly available, nor do their published descriptions furnish sufficient implementation details to enable reproducible evaluation.

We first consider fixed MACs percentages for presentation simplicity between subnetworks' methods for \dscnn S. \reds knapsack solution outperforms all three approaches, achieving higher accuracy. Moreover, we evaluate \reds against early exit linear classifiers~\citep{scardapane2020should} and magnitude-based width multiplier pruning on the \dscnn S and L architectures concerning MACs and test set accuracy (see \figref{fig:early_exits_ds_cnn_s_l_l1_comparison}). The subnetworks found by the \reds knapsack depth-wise formulation demonstrate superior performance over both methods, particularly within the domain of lower parameter configurations, specifically the S size.  

The last two plots on the right show \reds performance on mobile and IoT devices. We observe a difference of three orders of magnitude in the inference times between the two categories. \dnn models perform best thanks to the more optimized algorithm and libraries for matrix-matrix multiplication~\citep{goto2008anatomy}. All models show a linear relationship between the percentage of MACs and inference time. This empirical evidence validates MACs as a robust predictor of model latency, extending its applicability to mobile devices.  

On Arduino Nano 33 BLE Sense, TFLMicro framework extended to support \reds yields 38$\pm$1$\mu$s model adaptation time for a 2-layer fully-connected network, while the model inference times for the same network with 25\% and 50\% of MACs are 2'131$\pm$27$\mu$s and 4'548$\pm$13$\mu$s respectively. This highlights the efficiency of permutation-based approach adopted by \reds. The energy consumption of \reds inference as measured with the Power Profiler Kit (PPK2) on Nordic nRF52840 for \dscnn varies between 20mJ and 61mJ as shown in \tabref{tab:power_profiler_kit_benchmark}, whereas switching takes $<$0.01mJ.

\section{Conclusion}
\label{sec:conclusion}

This paper presents Resource-Efficient Deep Subnetworks (\reds), a novel approach to adapt deep neural networks to dynamic resource constraints typical for mobile and IoT devices. \reds formulates the subnetworks architectures search as an iterative knapsack problem, taking into account the dependencies between layers, MACs and peak memory usage. Moreover, \reds employs neuron permutation invariance to facilitate adaptation to variable resource availability without compromising the models' accuracy or adding runtime overhead. Notably, \reds ensures that the subnetworks' weights are stored in contiguous memory, enhancing cache optimization and compatibility with hardware-specific enhancements like vector instructions. Experimental evaluation on seven benchmark architectures demonstrates the effectiveness of \reds on mobile and IoT devices and superior performance of current pruning and neural architecture search state of the art methods. We perform a theoretical analysis of the knapsack solution space and prove the worst-case performance bounds for the two heuristic algorithms. 

There are several directions to extend this work. First, we have not tested the efficiency of \reds used explicitly as neural architecture search to discover neural network architectures for large language models and multimodal models. Second, the support of \reds optimization for caches in convolution models has not been explored and should be addressed in the future to support further advanced architectures. Third, scaling \reds solver to consider more constraints, \eg energy, is left for future work. We hope this work sparks interest in exploiting neural network symmetries to advance adaptive neural network deployment at the edge.

\section*{Acknowledgments}
The authors are grateful to Markus Gallacher for his support with the energy efficiency analysis of \reds. Christopher Hinterer and Julian Rudolf contributed to extending the TFL Micro framework to support \reds on Arduino Nano 33 BLE Sense.
This research was funded in part by the Austrian Science Fund (FWF) within the DENISE doctoral school (grant number DFH 5). The results presented in this paper were computed using the computational resources of the Vienna Scientific Cluster (VSC) and the HLR resources of the Zentralen Informatikdienstes of Graz University of Technology.

\bibliography{arxiv}

\newpage
\appendix
\def\dw{d}
\def\pf{f}
\def\pft{g}

\section{Iterative Knapsack Theory}
\label{appendix:iterative_knapsack}
    
In this section we prove that the order matters if we want to pack a knapsack iteratively. We are looking at the 2 stage iterative knapsack problem, where the items of a solution for capacity $c/2$ have to be a subset of the items of a solution for capacity $c$. 
We give our analysis and theoretical findings under the natural assumption that all items of the knapsack have weight $\leq c/2$. 
We first consider the bottom-up iterative knapsack heuristic and show that the quality of a worst-case solution is bounded by $\frac{2}{3} \cdot Opt$ and the bound is tight. We then analyze the top-down iterative knapsack heuristic and show that in this case a worst-case solution has a tight lower worst-case bound of $\frac{1}{2} \cdot Opt$.
Approximation results for the related incremental knapsack problem were given in~\citealt{DELLACROCE201926} and~\citealt{Faenza23}.
For a set of items $I$, let $P(I)$ ($W(I)$) denote the total profit (weight) of all items in $I$.
For a binary knapsack problem (see~\citealt{KePfPi04} for a general overview) we say that a \emph{split item} $I_s$ according to some ordering $O$ of the items and capacity $c$ exists, if there is an item $I_s$ with the following property: all the items $I_b^O$ that appear in $O$ before $I_s$ fulfill $W(I_b^O) < c$ and $W(I_b^O \cup I_s) > c$.

\paragraph{Bottom-up knapsack.}
Let $I(Opt_{c})$ denote the optimal solution set for a knapsack of capacity $c$.
Consider the following iterative heuristic $A_c$:
we first find an optimal solution of the knapsack with capacity $c/2$, then we fix the selected items $I(Opt_{c/2})$ and solve the knapsack defined on the remaining items and capacity $c - W(I(Opt_{c/2}))$. 
We denote this second set of items as $I(A_{c/2})$. 
Hence, the overall item set of this heuristic is given by $I(A_c) =I(Opt_{c/2}) \cup I(A_{c/2})$. 
Note that there may be $P(I(A_{c/2})) > P(I(Opt_{c/2}))$, since the corresponding knapsack capacity in the second step can be larger than $c/2$.

\begin{theorem}\label{th:kp1}
$A_c$ yields a worst case ratio of $\frac{2}{3}$, \ie $P(I(A_c)) \geq \frac{2}{3} \cdot P(I(Opt_{c}))$, if all items of the knapsack have a weight $\leq c/2$. This bound is tight.  
\end{theorem}
\begin{proof}
Let us arrange the items of $I(Opt_{c})$ in the following order $O$: 
we first take all the items from $I(Opt_{c}) \cap  I(Opt_{c/2})$ in an arbitrary order. Note that these are the items that are in both optimal solutions, \ie for both capacity $c$ and $c/2$. 
Then we take all the items that are not included in $I(A_{c/2})$ followed by the items of $I(Opt_{c}) \cap I(A_{c/2})$ (again in arbitrary order). 
Now we have two cases:

\medskip

\textit{Case 1:} 
There does not exist a split item in $I(Opt_{c})$ with respect to $O$ and capacity $c/2$. Hence $W(I(Opt_{c/2})) = c/2$. It is easy to see that in this case $A_c = Opt_{c}$.

\medskip

\textit{Case 2:} 
Let $I_s$ be the split item in $I(Opt_{c})$ with respect to $O$ and capacity $c/2$. 
In this case we get that the weight of all the items $I_b^O$ before $I_s$ as well as the weight of all the items $I_f^O$ that follow $I_s$ is smaller than $c/2$. 
It follows that $P(I(Opt_{c/2})) \geq P(I_b^O)$ and that $P(I(A_{c/2})) \geq P(I_f^O)$. 
Since all items have a weight $\leq c/2$ and by the fact that $I_s$ is not contained in $I(Opt_{c/2})$ we know that its profit is less or equal than the minimum of $P(I(A_{c/2}))$ and $P(I(Opt_{c/2}))$.
Therefore, it holds that $P(I_s) \leq \frac{1}{2}P(I(A_c))$. Hence we get:
\begin{eqnarray*}
P(I(Opt_c)) &= & P(I_b^O) + P(I_s) + P(I_f^O) \leq  P(I(A_c)) +  \frac{1}{2}P(I(A_c))\\
&=& \frac{3}{2} P(I(A_c)) 
\end{eqnarray*}

It remains to show the bound is tight. We introduce the following knapsack instance with four items and a large positive constant $P$.

\begin{center}
\begin{tabular}{|c|c|c|c|c| }
\hline
 item: & $1$ & $2$ &  $3$ & $4$  \\ 
 \hline
weight: & $c/3 + \epsilon$ & $c/3$ & $c/3$ & $c/3$\\  
 \hline
profit: & $P + \epsilon$ & $P$ & $P$ & $P$\\ 
 \hline
\end{tabular}
\end{center}

Here $I(Opt_{c/2})=\{1\}$ which only leaves space for one additional item for the larger capacity. 
Hence we get that $P(I(A_c)) = 2P + \epsilon$, whereas $P(I(Opt_c)) = 3P$. 
\end{proof}

\paragraph{Top-down knapsack.}
We now consider a heuristic $D_{c/2}$ consisting of an iterative top-down knapsack packing.
We first solve the knapsack with capacity $c$ to optimality, and then solve the knapsack problem defined only on the items $I(Opt_{c})$ with capacity $c/2$ to optimality. 
$I(D_{c/2})$ corresponds to the items in this second and smaller knapsack. 

\begin{theorem}
$D_{c/2}$ yields a worst case ratio of $\frac{1}{2}$, \ie $P(I(D_{c/2})) \geq \frac{1}{2} \cdot P(I(Opt_{c/2}))$ if all items of the knapsack have a weight $\leq c/2$. This bound is tight. 
\end{theorem}

\begin{proof}

Consider a knapsack of size $c$ with optimal solution set $I(Opt_c)$ and the knapsack problem with capacity $c/2$ defined on the restricted item set $I(Opt_{c})$ with solution set $I(D_{c/2})$. 
We will show that: $P(I(D_{c/2})) \geq \frac{1}{2} \cdot P(I(Opt_{c/2}))$.

\medskip
We first arrange the items of $I(Opt_c)$ in an ordering $O'$ such that they start with those items contained also in $I(Opt_{c/2})$. 
Then we identify the split item $I_s$ according to $O'$ for capacity $c/2$ and partition $I(Opt_c)$ into three parts.
$D_1 = I_b^{O'}$, $D_2 = I_s$ and $D_3$ contains all the remaining items. 
If no split item exists, we simply set $I_s=\emptyset$.
We now show that: 
\begin{equation}\label{eq:maxoptover2}
\max(P(D_1), P(D_2),P(D_3)) \geq \frac{P(I(Opt_{c/2}))}{2}
\end{equation}

Assuming that this is not the case, we would get that: 
$$\max(P(D_1), P(D_2),P(D_3)) < \frac{P(I(Opt_{c/2}))}{2}$$ 

\noindent 
This would imply
$$P(I(Opt_c)) =P(D_1)+P(D_2)+P(D_3) <  P(I(Opt_{c/2}))+P(D_3).$$
However, since $I(Opt_{c/2}) \cap D_3  = \emptyset$ and $W(I(Opt_{c/2})) \leq c/2$, $I(Opt_{c/2}) \cup D_3$ would constitute a feasible solution better than $I(Opt_c)$, which is a contradiction.
Thus, we have shown (\ref{eq:maxoptover2}).

For $i=1,\ldots,3$, there is $W(D_i) \leq c/2$ and all items in $D_i$ are available for $D_{c/2}$. 
Therefore, (\ref{eq:maxoptover2}) implies
$$P(I(D_{c/2})) \geq \max(P(D_1), P(D_2),P(D_3)) \geq \frac{1}{2} P(I(Opt_{c/2})).$$

It remains to show the bound is tight. We introduce the following knapsack instance with four items  and a large positive constant $P$.

\begin{center}
\begin{tabular}{|c|c|c|c|c| }
\hline
 Item: & $1$ & $2$ &  $3$ & $4$  \\ 
 \hline
 Weight: & $c/3$ & $c/3$ & $c/3$ & $c/2$\\  
 \hline
Profit: & $P + \epsilon$ & $P+ \epsilon$ & $P+ \epsilon$ & 2P\\ 
 \hline
\end{tabular}
\end{center}

\noindent Here $I(Opt_{c})=\{1,2,3\}$.
$D_{c/2}$ then selects one of these items and no more items fit into the knapsack. 
$Opt_{c/2}$ selects item $4$, which shows that the ratio of $\frac{1}{2}$ is tight. 
\end{proof}

Note that in case that we have instances, where the weight of certain items is greater that $c/2$, it is easy to construct instances with arbitrary bad ratios for both cases.

\section{Training Details}
\label{appendix:hyperparameters}

Table~\ref{tab:training_hyperparameters} summarizes the hyper-parameters used to train different networks. We refer to \citet{zhang2017hello} regarding the description of the network architectures adopted in this paper (referred to as \dnn, \cnn and \dscnn, sizes S and L). 

\begin{table}[h]
\footnotesize
    	\centering
    	\begin{tabular}{ccccc}
    		\toprule
    		\textbf{\textbf{Hyper-parameter}} & \textbf{DNN (S/L)}& \textbf{CNN (S/L)}& \textbf{\dscnn (S/L)} \\ 
    		\midrule
            Batch size       & \multicolumn{3}{c}{100} \\ 
            Training epochs  & 75  & 75 & 250 \\
            Loss function    & \multicolumn{3}{c}{Cross-entropy} \\ 
            Optimizer        & \multicolumn{3}{c}{Adam} & \\
            LR scheduler & \multicolumn{3}{c}{PiecewiseConstantDecay} \\
            LR    & $10^{-3}$, $10^{-4}$ & $5\cdot10^{-4}$, $10^{-4}$, $2\cdot10^{-5}$ & $10^{-3}$, $10^{-4}$ \\
            LR steps     & $1.5\cdot10^{4}$, & $2\cdot10^{4}$, $10^{4}$, $10^{4}$ & $10^{4}$, $10^{4}$, $10^{4}$ \\
                        & $3\cdot10^{3}$ & \\
    		\bottomrule
    	\end{tabular}
    	\caption{\textbf{Training hyper-parameters}.}
    	\label{tab:training_hyperparameters}
\end{table}

\section{Further \reds Performance Evaluation}
\label{appendix:further:eval}

\subsection{\reds performance on \dnn and \cnn architectures}

In addition to the results on \dscnn reported in the main paper, we show in \tabref{tab:finetuningvsfulltraining_dnn} and \tabref{tab:finetuningvsfulltraining_cnn} \reds performance on \dnn and \cnn architectures (with full fine-tuning) and compare to training model of each capacity from scratch and training \reds from scratch. Despite full fine-tuning, the results for S architecture show superior performance of the \bottomup heuristic over \topdown.

\begin{table*}[h]
\centering
\resizebox{\textwidth}{!}{
\begin{tabular}{c|cccc|cccc}
\toprule
\textbf{MACs}
& \multicolumn{4}{c|}{\textbf{Small (S) - Accuracy 83.82}} 
& \multicolumn{4}{c}{\textbf{Large (L) - Accuracy 86.87}} \\
& Scratch & Knapsack \bottomup & Knapsack \topdown & \reds training
& Scratch & Knapsack \bottomup & Knapsack \topdown & \reds training\\
\midrule
100\%                     & 84.30 $\pm0.11$  & 83.52 $\pm0.07$  & 82.80   $\pm0.16$ & 82.13 $\pm0.20$  & 86.54 $\pm0.24$ & 86.46  $\pm0.34$ & 86.25$\pm0.19$ & 85.06  $\pm0.19$ \\
75\%                      & 83.77 $\pm0.23$ & 82.29 $\pm0.35$  & 81.88  $\pm0.30$  & 81.23 $\pm0.21$  & 85.96 $\pm0.13$ & 86.38 $\pm0.65$ & 86.09$\pm0.03$ & 84.93  $\pm0.20$  \\
50\%                      & 80.91 $\pm0.11$ & 78.36 $\pm1.40$   & 78.59  $\pm0.24$ & 77.05 $\pm0.34$  & 85.24 $\pm0.35$ & 85.62 $\pm0.24$ & 85.58$\pm0.35$ & 84.08 $\pm0.22$ \\
25\%                      & 69.77 $\pm0.67$ & 64.42 $\pm1.99$  & 61.43  $\pm3.35$ & 63.69 $\pm3.26$  & 84.22 $\pm0.13$ & 82.61 $\pm0.61$ & 83.00   $\pm0.62$ & 82.03  $\pm0.59$ \\
\bottomrule
\end{tabular}}
\caption{Test set accuracy [\%] of training S and L fully-connected (\dnn) architectures taken from \citet{zhang2017hello}: training a network of each size from scratch ("Scratch"), conversion from a pre-trained network using two knapsack versions ("Knapsack BU" and "Knapsack TD"), and training \reds structure from scratch ("\reds training"). Reported results from three independent runs. The accuracy of each 100\,\% network reported in \cite{zhang2017hello} is listed in the header row.} 
\label{tab:finetuningvsfulltraining_dnn}
\end{table*}

\begin{table*}[h]
\centering
\resizebox{\textwidth}{!}{
\begin{tabular}{c|cccc|cccc}
\toprule
\textbf{MACs}
& \multicolumn{4}{c|}{\textbf{Small (S) - Accuracy 92.24}}    
& \multicolumn{4}{c}{\textbf{Large (L) - Accuracy 93.24}} \\  
& Scratch & Knapsack \bottomup & Knapsack \topdown & \reds training
& Scratch & Knapsack \bottomup & Knapsack \topdown & \reds training\\
\midrule
100\%                    & 91.10$\pm 0.23$  & 91.60$\pm 0.39$  & 91.20  $\pm 0.35$ & 88.89 $\pm0.26$ & 92.94$\pm 0.20$ & 92.83$\pm 0.26$  & 92.97$\pm0.15$ & 90.97 $\pm0.21$ \\
75\%                     & 90.40$\pm 0.27$  & 90.63$\pm 0.19$ & 90.20  $\pm 0.13$ & 87.64 $\pm0.46$ & 92.74$\pm 0.12$ & 92.74$\pm 0.23$ & 92.54 $\pm 0.07$ & 90.67 $\pm0.09$ \\
50\%                     & 89.07$\pm 0.24$ & 88.39$\pm 0.31$ & 88.39 $\pm 0.52$ & 85.99 $\pm0.19$ & 92.44$\pm0.30$ & 91.95$\pm 0.22$   & 91.93 $\pm0.28$ & 90.36 $\pm0.30$ \\
25\%                     & 82.57$\pm 0.41$ & 79.52$\pm 0.67$ & 79.28 $\pm 0.54$ & 79.25 $\pm0.40$ & 90.98$\pm0.60$ & 90.24$\pm 0.35$    & 90.31 $\pm0.03$ & 88.25 $\pm0.66$ \\
\bottomrule
\end{tabular}}
\caption{The same as in \tabref{tab:finetuningvsfulltraining_dnn} for S and L convolutional architectures (\cnn) from \citet{zhang2017hello}.} 
\label{tab:finetuningvsfulltraining_cnn}
\end{table*}

\begin{figure*}[h]
    \centering
    \includegraphics[width=0.245\linewidth]{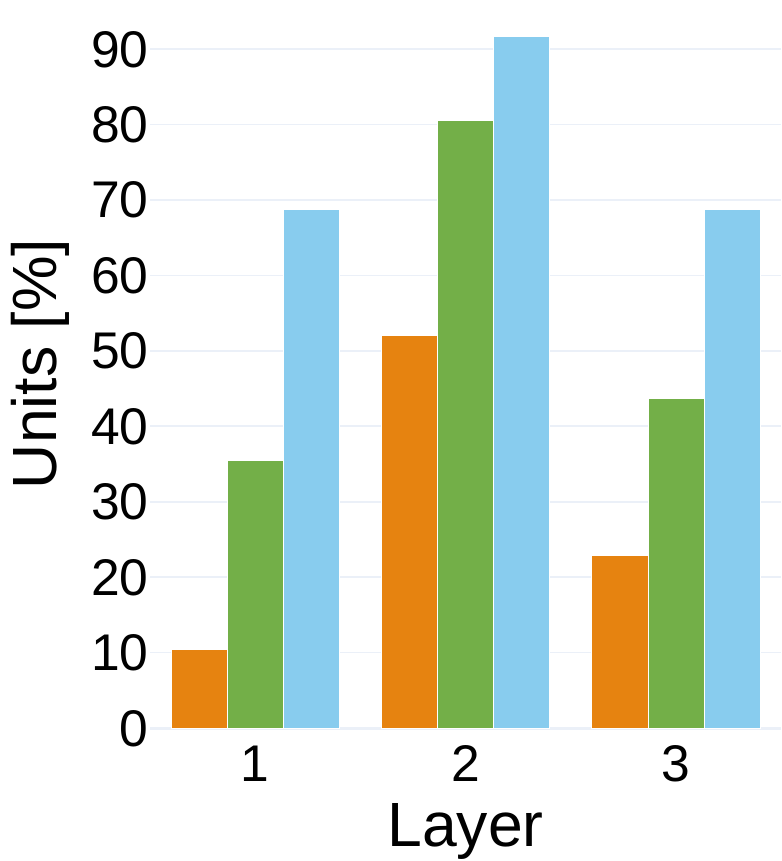}
    \includegraphics[width=0.245\linewidth]{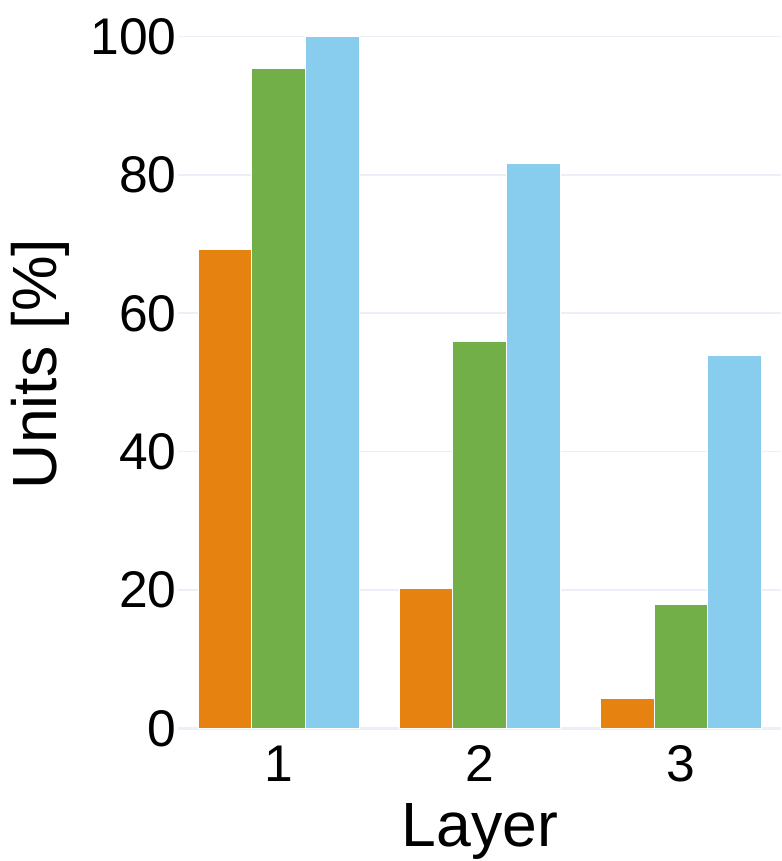}   
    \includegraphics[width=0.245\linewidth]{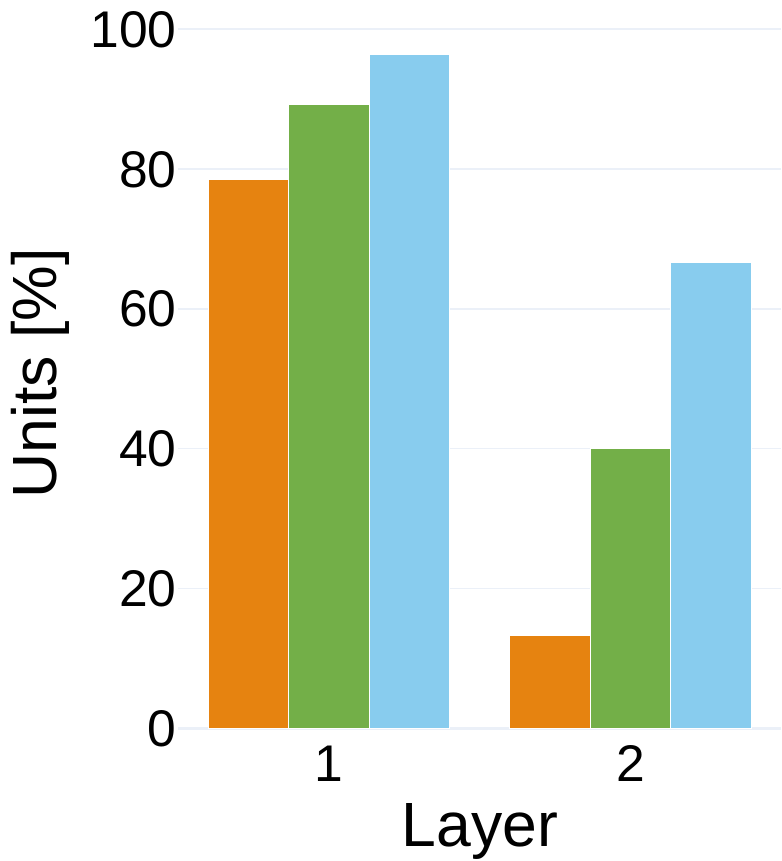}
    \includegraphics[width=0.245\linewidth]{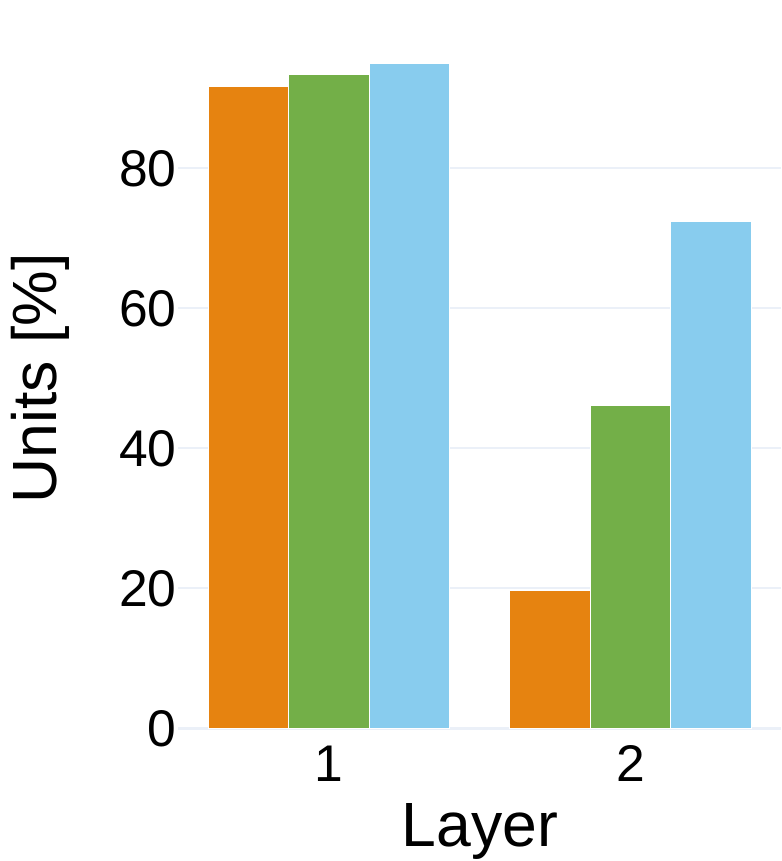}
    \caption{Analysis of the subnetwork architecture obtained by the knapsack \bottomup heuristics. From left to right: \dnn S, \dnn L, \cnn S and \cnn L on \googlespeech. The patterns as to which computational units constitute a child subnetwork are architecture-specific.}
    \label{fig:structural_pruning_models}
\end{figure*}

\figref{fig:structural_pruning_models} shows the impact of the architecture on \reds structure found by the knapsack \bottomup solver. We present the results for \dnn S, L and \cnn S, L from left to right, respectively.

\subsection{\reds performance with 10 nested subnetworks}
\begin{figure*}[h]
    \centering
    \includegraphics[width=0.28\linewidth]{figs/granularity/plotly_accuracy_sizeS_10subnetworks.pdf}
    \includegraphics[width=0.28\linewidth]{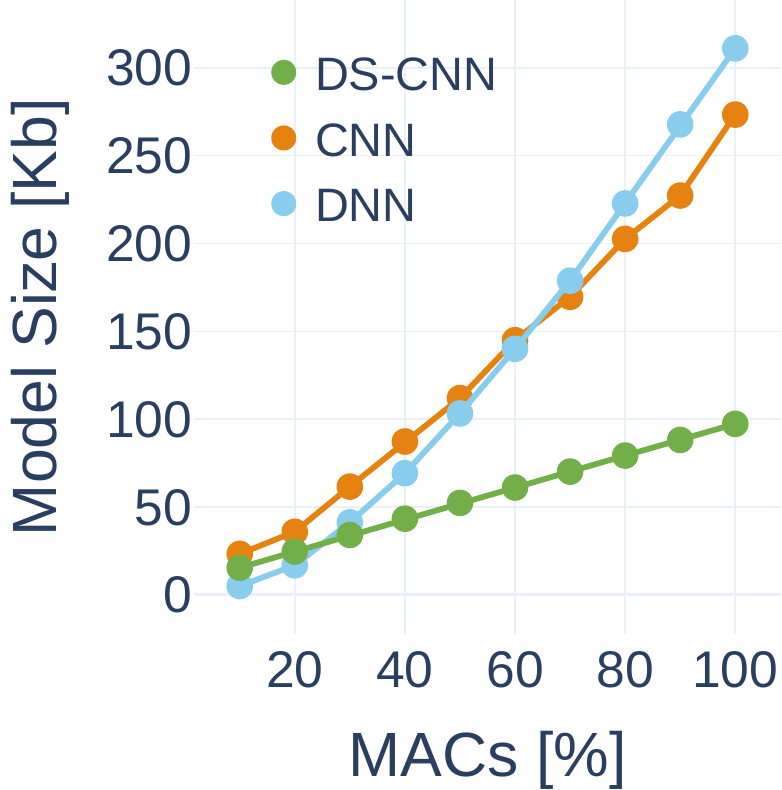}
    \includegraphics[width=0.28\linewidth]{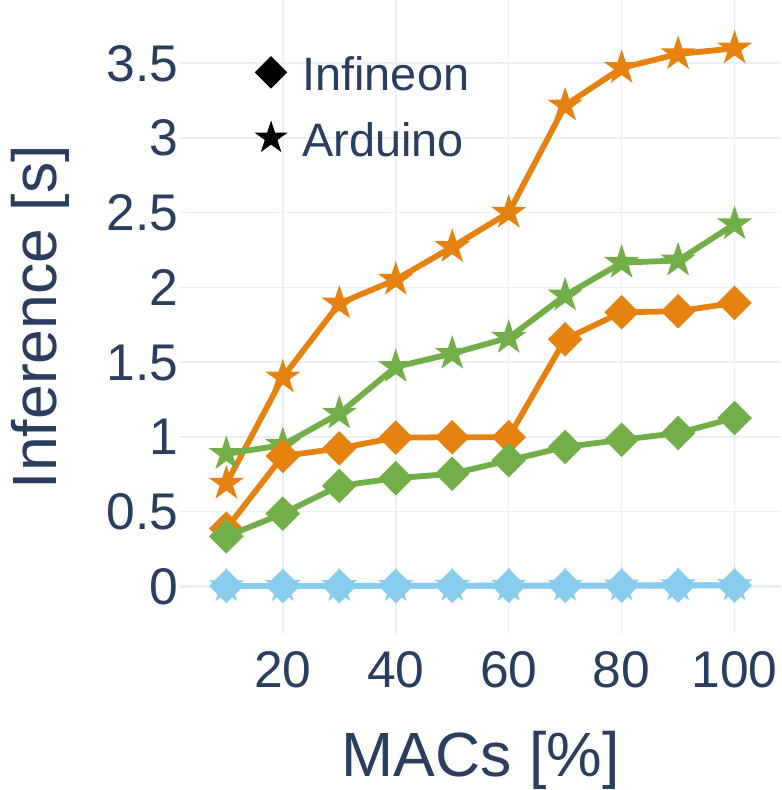}

    \includegraphics[width=0.28\linewidth]{figs/granularity/plotly_accuracy_sizeL_10subnetworks.pdf}
    \includegraphics[width=0.28\linewidth]{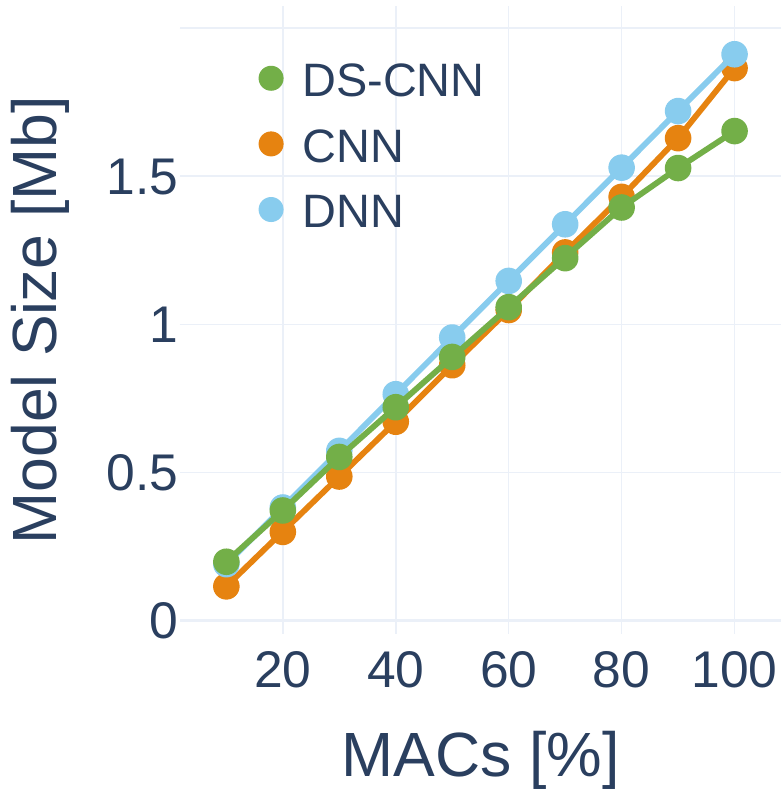}    
    \includegraphics[width=0.28\linewidth]{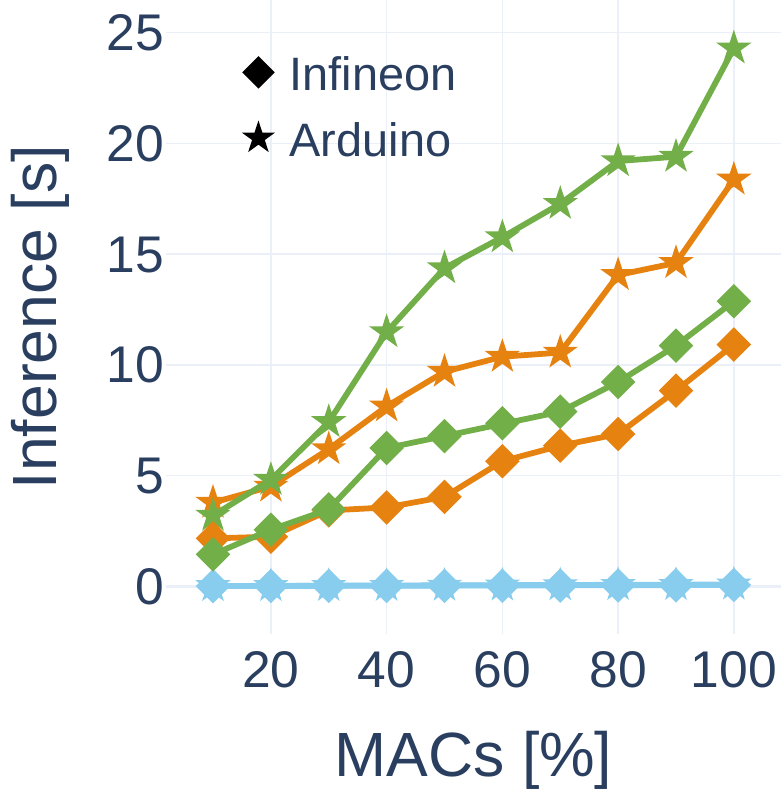}     
    \caption{\reds size S (top row) and L (bottom row) architectures analysis finetuned on Google Speech Commands~\citep{warden2019SC} with ten subnetworks. The plots from left to right show the subnetworks size, the subnetworks accuracy and the subnetworks inference time as a function of MAC percentage.
    }
    \label{fig:10subnetworks_size}
\end{figure*}

\tabref{tab:kws_10subnetworks} and \figref{fig:10subnetworks_size} show the performance of \dnn, \cnn and \dscnn on \googlespeech, when \reds structure comprises 10 subnetworks, compared to 4 subnetworks in the main paper. A larger number of subnetworks does not degrade model accuracy.

\begin{table*}[h]
\centering
	\begin{tabular}{c|cc|cc|cc}
    \toprule
    \textbf{MACs} &
    \multicolumn{2}{c}{\textbf{\dnn}} & 
    \multicolumn{2}{c}{\textbf{\cnn}} & 
    \multicolumn{2}{c}{\textbf{\dscnn}} \\
    & Small & Large &
    Small & Large &
    Small & Large \\
    \midrule
    100\%  & 83.07 $\pm 0.35$  & 86.19 $\pm 0.26$ & 91.16 $\pm 0.45$ & 93.1  $\pm 0.1 $ & 93.5  $\pm 0.15$ & 94.34 $\pm 0.07$ \\
    90\%   & 82.93 $\pm 0.4$   & 86.17 $\pm 0.36$ & 90.4  $\pm 0.47$ & 92.84 $\pm 0.27$ & 93.33 $\pm 0.11$ & 94.32 $\pm 0.1$  \\
    80\%   & 82.67 $\pm 0.53$  & 86.1  $\pm 0.08$ & 90.1  $\pm 0.07$ & 92.77 $\pm 0.06$ & 92.84 $\pm 0.15$ & 94.31 $\pm 0.1$  \\
    70\%   & 81.67 $\pm 0.53$  & 85.78 $\pm 0.11$ & 89.43 $\pm 0.41$ & 92.25 $\pm 0.47$ & 92.64 $\pm 0.24$ & 94.21 $\pm 0.14$ \\
    60\%   & 80.37 $\pm 0.57$  & 85.66 $\pm 0.25$ & 88.84 $\pm 0.25$ & 92.28 $\pm 0.11$ & 92.27 $\pm 0.31$ & 94.08 $\pm 0.03$ \\
    50\%   & 78.26 $\pm 0.53$  & 85.33 $\pm 0.09$ & 88.4  $\pm 0.2 $ & 92.03 $\pm 0.23$ & 91.04 $\pm 0.51$ & 93.97 $\pm 0.04$ \\
    40\%   & 75.37 $\pm 1.26$  & 84.8  $\pm 0.45$ & 85.76 $\pm 0.18$ & 91.78 $\pm 0.04$ & 89.42 $\pm 0.66$ & 93.83 $\pm 0.22$ \\
    30\%   & 67.76 $\pm 1.59$  & 82.66 $\pm 0.18$ & 81.92 $\pm 0.49$ & 90.52 $\pm 0.54$ & 87.63 $\pm 1.03$ & 93.59 $\pm 0.14$ \\
    20\%   & 52.36 $\pm 6.99$  & 80.19 $\pm 0.39$ & 73.74 $\pm 0.04$ & 88.61 $\pm 0.51$ & 84.14 $\pm 2.57$ & 93.35 $\pm 0.15$ \\
    10\%   & 23.93 $\pm 5.88$  & 50.7  $\pm 7.5 $ & 58.87 $\pm 5.27$ & 81.15 $\pm 1.39$ & 58.46 $\pm 3.35$ & 90.38 $\pm 0.17$ \\
    \bottomrule
    \end{tabular}
\caption{Test set accuracy [\%] from Small (S) and Large (L) pretrained fully-connected (\dnn), convolutional (\cnn), and depth-wise separable convolutional (\dscnn) networks taken from \citet{zhang2017hello}. For each pre-trained architecture, \reds can support ten subnetworks obtained from the Knapsack \bottomup formulation. The accuracies of the \dscnn and \cnn subnetworks do not degrade drastically until the lowest percentage of MACs considered. In contrast, the accuracies in the \dnn subnetworks show a more pronounced drop from 30\% MACs.} 

\label{tab:kws_10subnetworks}
\end{table*}

\clearpage
\subsection{\reds on \fmnist and \cifar}
The results in \tabref{tab:fashionmnist_analysis} and \tabref{tab:cifar10_analysis} show \reds performance using \dscnn architecture of size S on \fmnist and \cifar. \bottomup heuristic was used to obtain the results. \reds supports a different data domain without degrading the accuracy of the pre-trained model, reported in the header row. Compared to the state-of-the-art such as $\mu$NAS~\citep{liberis2021munas}, \reds demonstrates a faster architecture search time for both \fmnist and \cifar. In the former, \reds takes 19 minutes as opposed to 3 days; in the latter, \reds takes 90 minutes as opposed to 39 days while requiring less memory for model storage for both datasets. After finding and freezing the 25\% MACs subnetwork architecture, the \bottomup heuristic takes only a few seconds to find the other 50\% and 75\% MACs subnetworks architectures.  

\begin{table*}[h]
\centering
    \begin{tabular}{c|c|c|c}
    \toprule
    \textbf{MACs} & \textbf{Acc (\%) - Pre-trained  90.59} & \textbf{Model Size (Kb)} & \textbf{Time Taken (m)} \\
    \midrule
    100\% & 91.6 $\pm 0.2$  & 128.54 & -- \\
    75\%  & 91.51 $\pm 0.28$   & 107.73 &  1.58 [s] \\ %
    50\%  & 90.75    $\pm 0.32$   & 87.4 &  5.83 [s] \\ %
    25\%  & 89.22 $\pm 0.45$  & 66.63 & 19 [m] \\
    \bottomrule
    \end{tabular}
    \caption{Analysis of the \bottomup knapsack subnetworks obtained from a depth-wise separable convolutional (\dscnn) S network, pre-trained on \fmnist~\citep{xiao2017fashion}. 
    \reds supports a different data domain without degrading the accuracy of the pre-trained model, reported in the header row.
    }
    \label{tab:fashionmnist_analysis}
\end{table*}

\begin{table*}[h]
\centering
    \begin{tabular}{c|c|c|c}
    \toprule
    \textbf{MACs} & \textbf{Acc (\%) - Pre-trained 79.36} & \textbf{Model Size (Kb) } & \textbf{Time Taken} \\
    \midrule
    100\% & 81.07 $\pm 0.71$  & 128.54 & -- \\
    75\%  & 80.17  $\pm 0.69$  & 109.41 &  2.89  [s] \\   %
    50\%  & 76.72  $\pm 1.37$  & 88.01 &  10.59  [s] \\ %
    25\%  & 68.66 $\pm 1.65$  & 69.63 &  90 [m] \\ 
    \bottomrule
    \end{tabular}
    \caption{The same evaluation as in \tabref{tab:fashionmnist_analysis} for \cifar~\citep{cifar100}. 
    }
    \label{tab:cifar10_analysis}
\end{table*}

\subsection{\reds energy efficiency}
\tabref{tab:power_profiler_kit_benchmark} reports the energy consumption of inference time by different network architectures measured by Power Profiler Kit (PPK2) on Nordic nRF52840 (Arduino Nano 33 BLE Sense). In comparison, the model adaptation time takes less than 0.01\,mJ. 

\begin{table*}[h]
\centering
	\begin{tabular}{c|c|c|c}
    \toprule
    \textbf{MACs} & 
    \textbf{\dnn} & 
    \textbf{\cnn} & 
    \textbf{\dscnn} \\
          & [mJ] & [mJ] & [mJ] \\
    \midrule
    100\% & 0.18 & 90.61 & 61 \\
    75\%  & 0.15 & 86.01 & 44.03 \\
    50\%  & 0.07 & 50.3 & 33.81 \\
    25\%  & 0.05 & 44.81 & 20.44 \\
    \bottomrule
    \end{tabular}
    \caption{Energy consumption of \reds size S architectures measured by the Power Profiler Kit (PPK2) on Nordic nRF52840. The results are obtained by performing an inference pass for each subnetwork model and recording the inference current.} 
    \label{tab:power_profiler_kit_benchmark}
\end{table*}

\end{document}